\documentclass[10pt,twocolumn,letterpaper]{article}

\usepackage{cvpr}              %

\usepackage{amsmath,amsfonts,amsthm,amssymb}
\usepackage{mathtools}
\usepackage{bm}
\usepackage{nicefrac}
\usepackage{microtype}
\usepackage{lipsum}

\usepackage{color,xcolor}
\usepackage{epsfig}
\usepackage{graphicx}

\usepackage{adjustbox}
\usepackage{array}
\usepackage{booktabs}
\usepackage{colortbl}
\usepackage{wrapfig}
\usepackage{hhline}
\usepackage{multirow}
\usepackage{subcaption}
\usepackage[size=small]{caption}
\usepackage{float}

\usepackage{changepage}
\usepackage{extramarks}
\usepackage{fancyhdr}
\usepackage{lastpage}
\usepackage{setspace}
\usepackage{soul}
\usepackage{xspace}

\usepackage{url}

\usepackage{algpseudocode}
\usepackage{algorithmicx}
\usepackage[ruled]{algorithm2e}
\usepackage{enumerate}
\usepackage{enumitem}  %
\usepackage{makecell}
\usepackage{pifont}
\usepackage{titlecaps}
\usepackage[accsupp]{axessibility}
\usepackage{framed}

\definecolor{formalbar}{rgb}{0.290,0.325,0.337}
\definecolor{formalshade}{rgb}{0.925,0.941,0.976}

\newcolumntype{L}[1]{>{\raggedright\let\newline\\\arraybackslash\hspace{0pt}}m{#1}}
\newcolumntype{C}[1]{>{\centering\let\newline\\\arraybackslash\hspace{0pt}}m{#1}}
\newcolumntype{R}[1]{>{\raggedleft\let\newline\\\arraybackslash\hspace{0pt}}m{#1}}

\newcommand{\sect}[1]{Section~{#1}}

\newcommand{\eqn}[1]{Equation~{#1}}
\newcommand{\fig}[1]{Fig.~{#1}}
\newcommand{\tbl}[1]{Table~{#1}}

\newcommand{\ignore}[1]{}

\DeclareMathAlphabet{\mathbfit}{OML}{cmm}{b}{it}

\makeatletter
\DeclareRobustCommand\onedot{\futurelet\@let@token\@onedot}
\def\@onedot{\ifx\@let@token.\else.\null\fi\xspace}

\def\eg{e.g\onedot}

\makeatother

\definecolor{MyDarkBlue}{rgb}{0,0.08,1}
\definecolor{MyAqua}{rgb}{0,0.7,0.7}
\definecolor{MyDarkGreen}{rgb}{0.02,0.6,0.02}
\definecolor{MyDarkRed}{rgb}{0.8,0.02,0.02}
\definecolor{MyDarkOrange}{rgb}{0.40,0.2,0.02}
\definecolor{MyPurple}{RGB}{111,0,255}
\definecolor{MyRed}{rgb}{1.0,0.0,0.0}
\definecolor{MyGold}{rgb}{0.75,0.6,0.12}
\definecolor{MyDarkgray}{rgb}{0.66, 0.66, 0.66}

\definecolor{Cardinal}{rgb}{0.549,0.082,0.082}

\newif\ifdrafting
\draftingtrue %
\ifdrafting
    \newcommand{\jw}[1]{\textcolor{MyDarkGreen}{[Jiajun: #1]}}
    \newcommand{\ky}[1]{\textcolor{Cardinal}{[Koven: #1]}}
    \newcommand{\ds}[1]{{\leavevmode\color[rgb]{0.8,0.2,0}[Deqing: #1]}}
    \newcommand{\cih}[1]{{\textcolor{MyAqua}{[Charles: #1]}}}
    
    \newcommand{\samirag}[1]{\textcolor{blue}{[Samir: #1]}}

\else
    \newcommand{\ds}[1]{}
    \newcommand{\cih}[1]{}
    \newcommand{\jw}[1]{}
    \newcommand{\ky}[1]{}
    \newcommand{\samirag}[1]{}
\fi

\SetKwFor{For}{for }{}{}

\newcommand{\modelfull}{WonderJourney\xspace}
\newcommand{\model}{WonderJourney\xspace}
\newcommand{\llmpart}{scene description generation}
\newcommand{\visionpart}{visual scene generation}
\newcommand{\vlmpart}{visual validation}

\newcommand{\problem}{perpetual 3D scene generation\xspace}

\DeclareMathOperator{\R}{\mathbb{R}}
\DeclareMathOperator{\Loss}{\mathcal{L}}

\newcommand{\myparagraph}[1]{\vspace{0.1cm}\noindent\textbf{#1}}

\newcommand{\aftersec}{\vspace{-.4em}}
\newcommand{\aftersubsec}{\vspace{-.2em}}
\newcommand{\aftertab}{\vspace{-1em}}
\newcommand{\afterfig}{\vspace{-1.25em}}
\newcommand{\aroundeqn}{}

\definecolor{cvprblue}{rgb}{0.21,0.49,0.74}
\usepackage[pagebackref,breaklinks,colorlinks,citecolor=cvprblue]{hyperref}

\def\paperID{1976} %
\def\confName{CVPR}
\def\confYear{2024}

\title{\modelfull: Going from Anywhere to Everywhere}

\newcommand{\authorhsfirst}{\hspace{5mm}}
\newcommand{\authorhssceond}{\hspace{2mm}}
\author{Hong-Xing Yu\textsuperscript{1} \authorhsfirst
Haoyi Duan\textsuperscript{1} \authorhsfirst
Junhwa Hur\textsuperscript{2} \authorhsfirst
Kyle Sargent\textsuperscript{1} \authorhsfirst
Michael Rubinstein\textsuperscript{2} \authorhsfirst
\vspace{0.15cm}\\
William T. Freeman\textsuperscript{2} \authorhssceond
Forrester Cole\textsuperscript{2} \authorhssceond
Deqing Sun\textsuperscript{2} \authorhssceond
Noah Snavely\textsuperscript{2} \authorhssceond
Jiajun Wu\textsuperscript{1} \authorhssceond
Charles Herrmann\textsuperscript{2} \authorhssceond
\vspace{0.25cm}\\
\textsuperscript{1}Stanford University  \hspace{10mm}
\textsuperscript{2}Google Research
}

\begin{document}

\twocolumn[{%
\renewcommand\twocolumn[1][]{#1}%
\maketitle
\begin{center}
    \centering
    \captionsetup{type=figure}
    \vspace{-0.7cm}
    \includegraphics[width=1\textwidth]{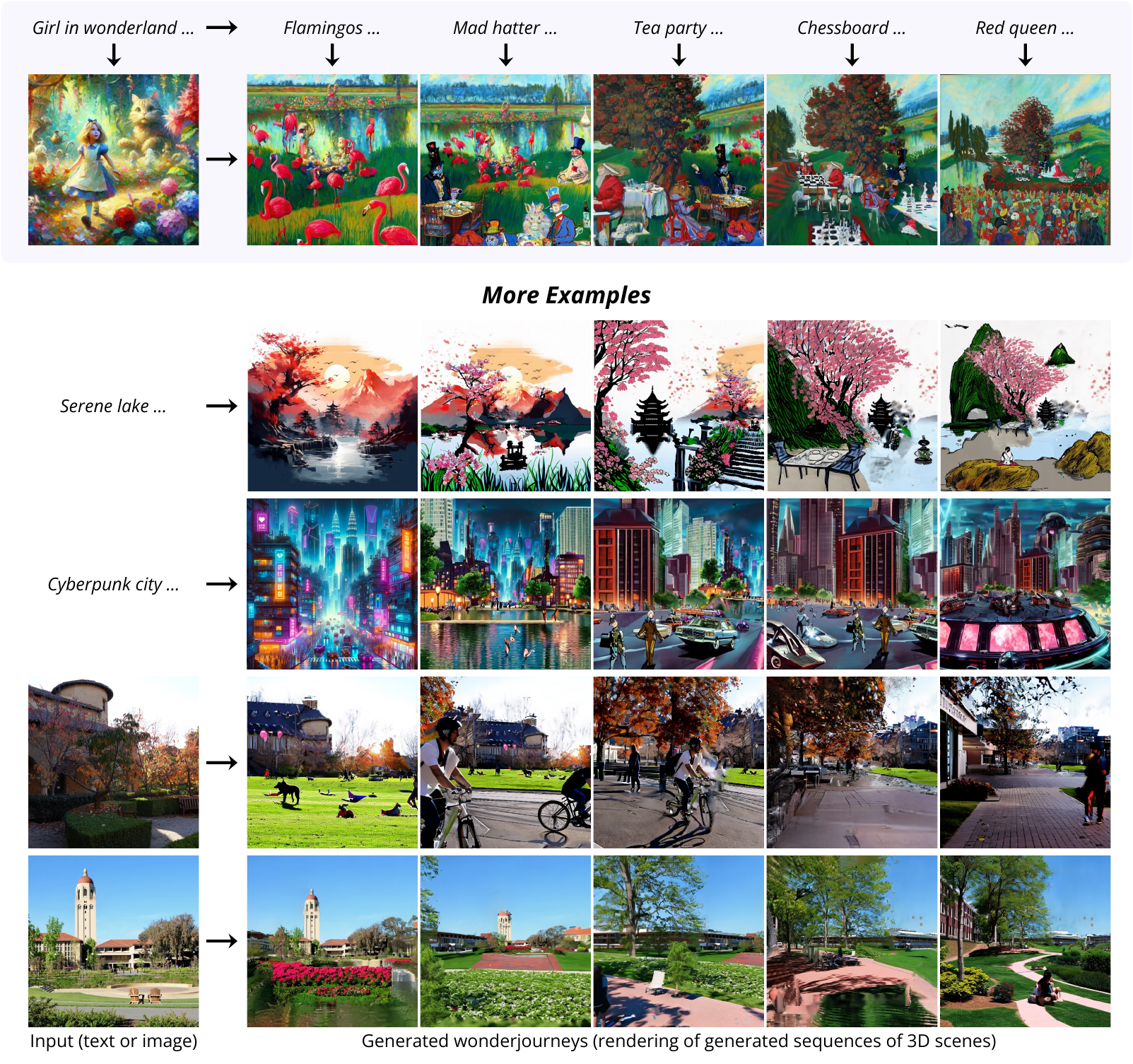}
    \captionof{figure}{We propose \textbf{\model}---generating a sequence of diverse yet coherent 3D scenes from a text description or an arbitrary image such as photos or generated art (``from anywhere''). \model can generate various journeys (which we refer to as ``wonderjourneys'') for a fixed input, potentially ending ``everywhere'' (\fig{\ref{fig:diverse_results}}). We show rendered images along the generated sequence of 3D scenes. \textbf{We strongly encourage the reader to see video examples at \url{https://kovenyu.com/WonderJourney/}.}
    }
    \label{fig:teaser}
    \vspace{-0.6cm}
\end{center}%
}]

\begin{abstract}
\aftersec
We introduce \modelfull, a modular framework for \problem. Unlike prior work on view generation that focuses on a single type of scenes, we start at any user-provided location (by a text description or an image), and generate a journey through a long sequence of diverse yet coherently connected 3D scenes. 
We leverage an LLM to generate textual descriptions of the scenes in this journey, a text-driven point cloud generation pipeline to make a compelling and coherent sequence of 3D scenes, and a large VLM to verify the generated scenes. We show compelling, diverse visual results across various scene types and styles, forming imaginary ``wonderjourneys''. 
Project website: \url{https://kovenyu.com/WonderJourney/}.
\end{abstract}
\vspace{-10pt}
\begin{flushright}
``No, no! The adventures first, explanations take such a dreadful time.'' -- \emph{Alice's Adventures in Wonderland}
\end{flushright}

\section{Introduction}
\aftersec
In %
``Alice's Adventures in Wonderland'', the story begins with Alice falling down the rabbit hole and emerging into a strange and captivating Wonderland. In her journey through this wonderland, Alice encounters many 
curious characters
such as the Cheshire Cat and the Mad Hatter,
and peculiar scenarios, such as the tea party and the rose garden -- eventually ending at the royal palace. These characters and settings combine to form a compelling world that has fascinated countless readers over the years. In this paper, we follow in this creative tradition and explore how modern computer vision and AI models can similarly generate such interesting and varied visual worlds, which users can journey through, much like Alice did in her own adventures in Wonderland. 

Toward this goal, we introduce the problem of \emph{\problem}. Unlike previous work on perpetual view generation~\citep{liu2021infinite,li2022infinitenature} that only generates a single type of scene, such as nature photos, our objective is to synthesize a series of diverse 3D scenes starting at an arbitrary location specified via a single image or language description. The generated 3D scenes should be \emph{coherently connected} along a long-range camera trajectory, traveling through various plausible places. The generated 3D scenes allow rendering a fly-through video through a long series of diverse scenes to simulate the visual experience of a journey in an imaginary ``wonderland''. We show examples in \fig{\ref{fig:teaser}}. 

The main challenges of \problem center around generating \emph{diverse yet plausible} scene elements. These scene elements need to support the formation of a path through coherently connected 3D scenes. They should include various objects, backgrounds, and layouts that fit in observed scenes and transit naturally to the next, yet unobserved, scenes. This generation process can be broken down into determining what objects to generate for a given scene; where to generate these objects; and how these scenes connect to each other geometrically. Determining what elements to generate calls for semantic understanding of a scene (e.g., a lion might not be a great fit for a kitchen); determining where to generate them calls for common sense regarding the visual world (e.g., a lion should not be floating in the sky); further, generating these elements in a new connected scene requires geometric understanding (e.g., occlusion and disocclusion, parallax, and appropriate spatial layouts). Notably, these challenges are absent in prior work on perpetual view generation, as they tend to focus on a single domain~\citep{liu2021infinite} or a single scene~\citep{fridman2023scenescape}.

Our proposed solution, \modelfull, addresses each of the above challenges in \problem with its own module. \model leverages an LLM to generate a long series of scene descriptions, followed by a text-driven visual scene generation module to produce a series of colored point clouds to represent the connected 3D scenes. 
Here, the LLM provides commonsense and semantic reasoning; the vision module provides visual and geometric understanding and the appropriate 3D effects. In addition, we leverage a vision-language model (VLM) to verify the generation and launch a re-generation when it detects undesired visual effects. Our framework is modular. Each of the modules can be implemented by the best pretrained models, allowing us to leverage the latest and future advancements in the rapidly growing vision and language research.
Our main contributions are as follows:
\vspace{0.1cm}
\begin{itemize}
    \item We study \problem and propose \model, a modular framework that decomposes this problem into core components: an LLM to generate scene descriptions, a text-driven visual module to generate the coherent 3D scenes, and a VLM to verify the generated scenes.
    \vspace{0.05cm}
    \item We design a visual scene generation module that leverages off-the-shelf text-to-image and depth estimation models to generate coherent 3D point clouds. Our module handles boundary depth inaccuracy, sky depth inaccuracy, and dis-occlusion unawareness.
    \vspace{0.05cm}
    \item We show compelling visual results and compare \model with SceneScape~\citep{fridman2023scenescape} and InfiniteNature-Zero~\citep{li2022infinitenature} in a user study, which shows that \model produces interesting and varied journeys. %
    
\end{itemize}

\begin{figure*}
    \centering
    \includegraphics[width=\textwidth]{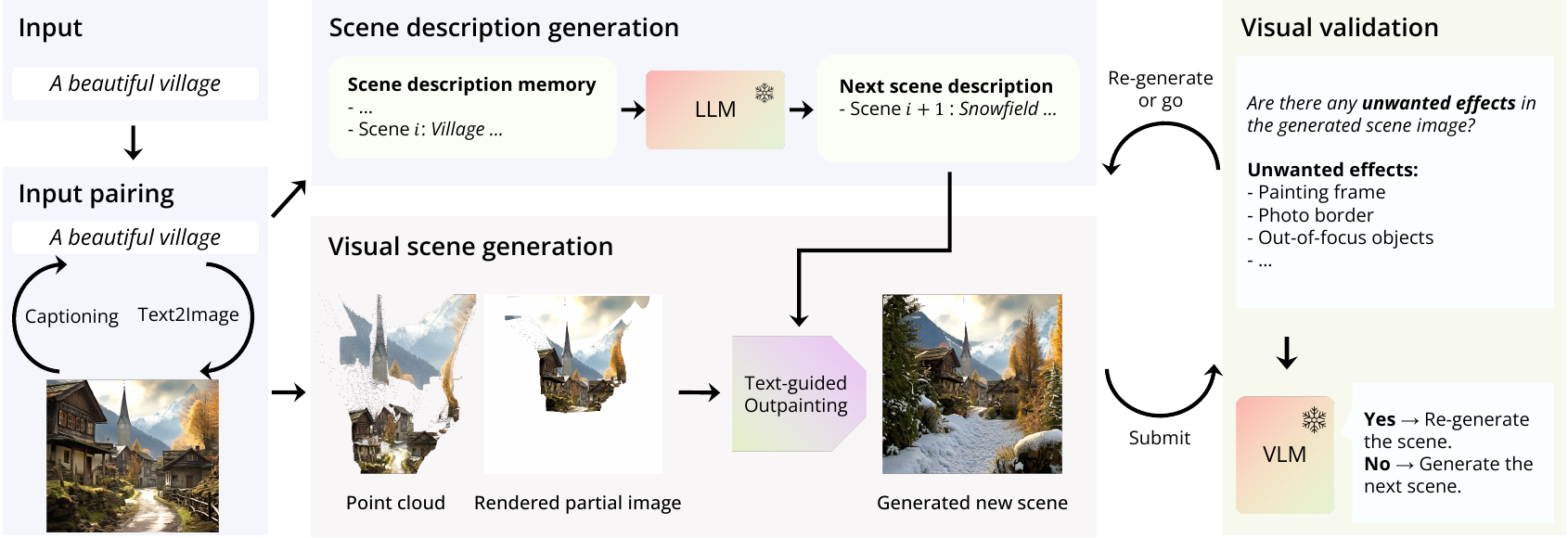}
    \vspace{-0.6cm}
    \caption{\textbf{The proposed \modelfull framework and  workflow across modules}. Our  modular design does not require any training, allowing easy future improvements from the quick advances in vision and language models.}
    \label{fig:overview}
    \afterfig
\end{figure*}
\section{Related Work}
\aftersec

\myparagraph{Perpetual view generation.}
The seminal work on perpetual view generation is Infinite Images \cite{kaneva2010infinite} which synthesizes the effect of navigating a 3D world by stitching and rendering images according to camera motions.
Later works, such as Infinite Nature~\cite{liu2021infinite} and InfiniteNature-Zero~\cite{li2022infinitenature}, learn to auto-regressively generate next view based on the current view. 
Follow-up works improve global 3D consistency~\citep{chai2023persistent} and visual quality~\cite{cai2022diffdreamer}. %
A recent work, SceneScape~\cite{fridman2023scenescape}, explores text-driven perpetual view generation by gradually constructing a single cave-like scene represented by a lengthy mesh.
While these methods study perpetual generation, they are restricted to a single domain such as nature photos~\cite{liu2021infinite,li2022infinitenature} or a single scene~\cite{fridman2023scenescape}.

\myparagraph{3D scene generation.}
Considerable progress has recently been made in text-to-3D or image-to-3D generation, many of which focus on objects without background~\cite{poole2022dreamfusion, raj2023dreambooth3d,liu2023zero,melas2023realfusion, lin2023magic3d, li20223ddesigner, cheng2023sdfusion}. 
These works typically leverage a 2D image prior from an image generation model, e.g., an image diffusion model~\cite{rombach2022high}, and then build a 3D representation, such as a NeRF~\cite{wang2021nerf}, by distilling the supervision of the 2D image generation model~\citep{poole2022dreamfusion}.
Other works on 3D object generation focus on learning a 3D generative model directly from 2D images~\citep{chan2022efficient, chan2021pi, gu2021stylenerf, hao2021gancraft, nguyen2019hologan, niemeyer2021giraffe, or2022stylesdf, schwarz2022voxgraf, sargent2023vq3d}. 

Several works also focus on generating a single 3D scene with background~\cite{devries2021unconstrained,bautista2022gaudi}. 
For example,
Text2Room~\cite{hollein2023text2room} generates a room-scale 3D scene from a single text prompt, using textured 3D meshes for their scene representation.
Other relevant works have focused on generating (sometimes called ``reconstructing'') a scene from limited observations, such as a single image. Long-Term Photometric Consistent NVS~\cite{yu2023long} generates single scenes from a source image by auto-regressively generating with a conditional diffusion model. 
GeNVS~\cite{chan2023genvs} and Diffusion with Forward Models~\cite{tewari2023diffusion} use an intermediate 3D representation but are trained and evaluated on each scene separately. ZeroNVS~\cite{sargent2023zeronvs} synthesizes a NeRF of a scene from a single image.
Their work focuses on generating a single scene while ours targets generating a coherently connected sequence of diverse scenes.

\myparagraph{Text-guided video generation.}
The idea of scene generation has also been explored in video generation. Several concurrent works like TaleCrafter~\cite{gong2023talecrafter} and others~\cite{liu2023intelligent,huang2023free,lin2023videodirectorgpt} also discuss the task of creating a series of videos which follow an LLM-generated story or script. However, all these works use different scenes as different clips in a video, resulting in hard cuts, while our \problem aims at generating sequences of coherently connected scenes.

\ignore{
\myparagraph{Depth and Segmentation Estimation.}
Though our paper is about \problem, we make extensive use of the estimation literature

- Need to put some of the papers that we use here e.g. MIDAS, Zoe-Depth, SAM, and discuss that despite the progress there are still failures.
}

\begin{figure*}
    \includegraphics[width=\textwidth]{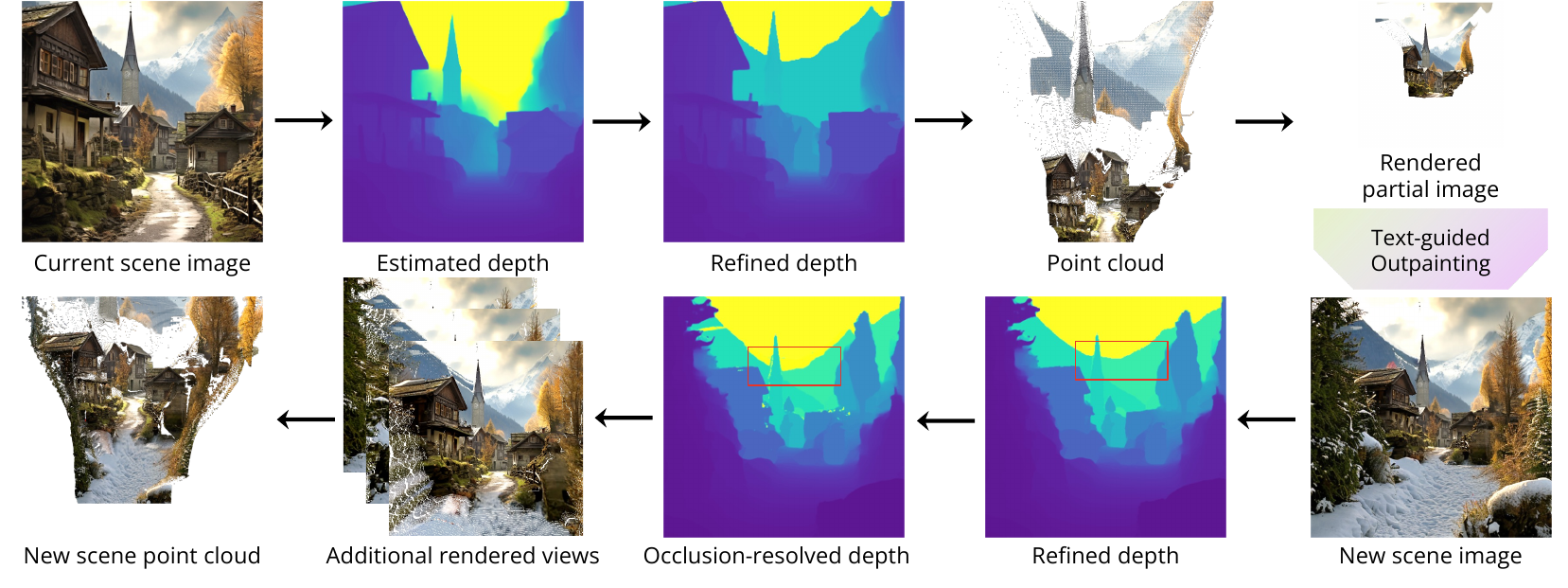}
    \vspace{-.8cm}
    \caption{\textbf{The visual scene generation module}. Each arrow represents a parametric vision model (e.g., a depth estimator) or an operation (e.g., rendering). Our fully modular design easily benefits from advances in the corresponding research topics.}
    \label{fig:vision_part}
    \afterfig
\end{figure*}
\section{Approach}
\aftersec

Our goal is to generate a potentially endless sequence of diverse yet coherently connected 3D scenes, which requires both geometric understanding of 3D scenes and visual common sense and semantic understanding of scene structures.
To tackle this challenging task, we propose \modelfull. The main idea is to generate a text description of the visual elements the next scene would contain, and then employ a text-guided visual generation module to make that 3D scene. 

\model is a modular framework that decomposes this task into \llmpart, \visionpart, and \vlmpart, as in \fig{\ref{fig:overview}}. Given an input image or text, we first pair it with the other modality by generating the image with a text-to-image model or by generating the description with a Vision Language Model (VLM). Then we generate the next-scene description by a Large Language Model (LLM). A visual scene generation module takes in the next-scene description and the current scene image to generate the next 3D scene represented by a colored point cloud. This generation process is then checked by a VLM to make sure there is no unwanted effects, or it gets regenerated. We note that our framework is modular such that each module can be implemented with the latest (pretrained) models and thus can easily leverage the quick advances of large language and vision models.

\subsection{Scene description generation}
\aftersubsec

We propose an auto-regressive scene description generation process, i.e., the \llmpart~ takes a set of past and current scene descriptions as input and predicts the subsequent scene description:
\aroundeqn
\begin{align}
    \mathcal{S}_{i+1} & = g_\text{description}(\mathcal{J}, \mathcal{M}_i),
\end{align}
\aroundeqn
where $\mathcal{S}_{i}$ denotes the $i^\text{th}$ scene description, and $g_\text{description}$ denotes the scene description generator which is implemented by an LLM that takes two inputs: 
1) the task specification $\mathcal{J}$ = \textit{``You are an intelligent scene generator. Please generate 3 most common entities in the next scene, along with a brief background description.''}
; and 2) the scene description memory $\mathcal{M}_{i} = \{\mathcal{S}_0, \mathcal{S}_1, \cdots, \mathcal{S}_i\}$ which is a collection of past and current scenes. The latest description memory is:
\aroundeqn
\begin{align}
    \mathcal{M}_{i+1} & = \mathcal{M}_i \cup \{\mathcal{S}_{i+1}\}.
    \label{eqn:memory}
\end{align}
\aroundeqn
We define the scene description $\mathcal{S}_i \!=\! \{ S, O_i, B_i \}$, which consists of a style $S$ that is kept consistent across scenes, objects in the scene $O_i$, and a concise caption $B_i$ describing the background of the scene. We allow the LLM to output natural language descriptions, and then use a lexical category filter to process the raw text of $O_i$ and $B_i$ such that we only keep nouns for entities and adjectives for attributes. Empirically this generates more coherently connected scenes compared to requiring the LLM to directly output this structured description.

\ignore{
\ky{Decide if we really want this empirical note to appear here. It seems to be a bit distracting to me. I prefer leaving this to supplement as a fun fact in 2023.} In our explorations, we empirically find potential disruptions induced by the inclusion of uncommon objects within textual descriptions, a factor that can detrimentally impact the 3D spatial quality quality of the generated scenes. For instance, an uncommon object "phone booth" has a high likelihood of being generated, leading to its placement in undesired and nonsensical locations, such as the middle of a road or floating in the sky. To address this, we propose instructing the LLM to prioritize providing "most common" objects, as mentioned in $\mathcal{J}$. 
}

\subsection{Visual scene generation}
\aftersubsec
Since we want the generated next scene to be coherent with past scenes geometrically and semantically, we formulate our \visionpart~ as a conditional generation problem, taking both the next-scene description and the 3D representation of the current scene as conditions: 
\aroundeqn
\begin{align}
    \mathcal{P}_{i+1} = g_\text{visual}(I_i, \mathcal{S}_{i+1}),
\label{eqn:visual}
\end{align}
\aroundeqn
where $\mathcal{P}_{i}$ denotes a colored point cloud that represents the next 3D scene, and $I_i$ denotes the image of current scene. The visual scene generator $g_\text{visual}$ consists of learning-free operations such as perspective unprojection and rendering, as well as components that use parametric (pretrained) vision models, including a depth estimator, a segmentation-based depth refiner, and a text-conditioned image outpainter. We show an illustration in \fig{\ref{fig:vision_part}}.

\myparagraph{Lifting image to point cloud.}\label{sec:imagetopointcloud}
Given the current scene represented by an image $I_i$, we lift it to 3D by estimating depth and unproject it with a pinhole camera model. We use MIDAS v3.1 \cite{Ranftl2022}, one of the state-of-the-art depth estimators, in our experiments. However, we find that existing monocular depth estimators share two common issues. First, depth discontinuity is not well modeled, witnessed by previous work~\citep{tosi2021smd,aleotti2021neural,miangoleh2021boosting}, resulting in overly smooth depth edges across object boundaries. Second, the depth of the sky is always underestimated, also observed by previous work~\citep{liu2021infinite,li2022infinitenature}. To address these two issues, we introduce a depth refinement process that leverages pixel grouping segments and sky segmentation.

\begin{figure*}
\centering
\includegraphics[width=0.95\textwidth]{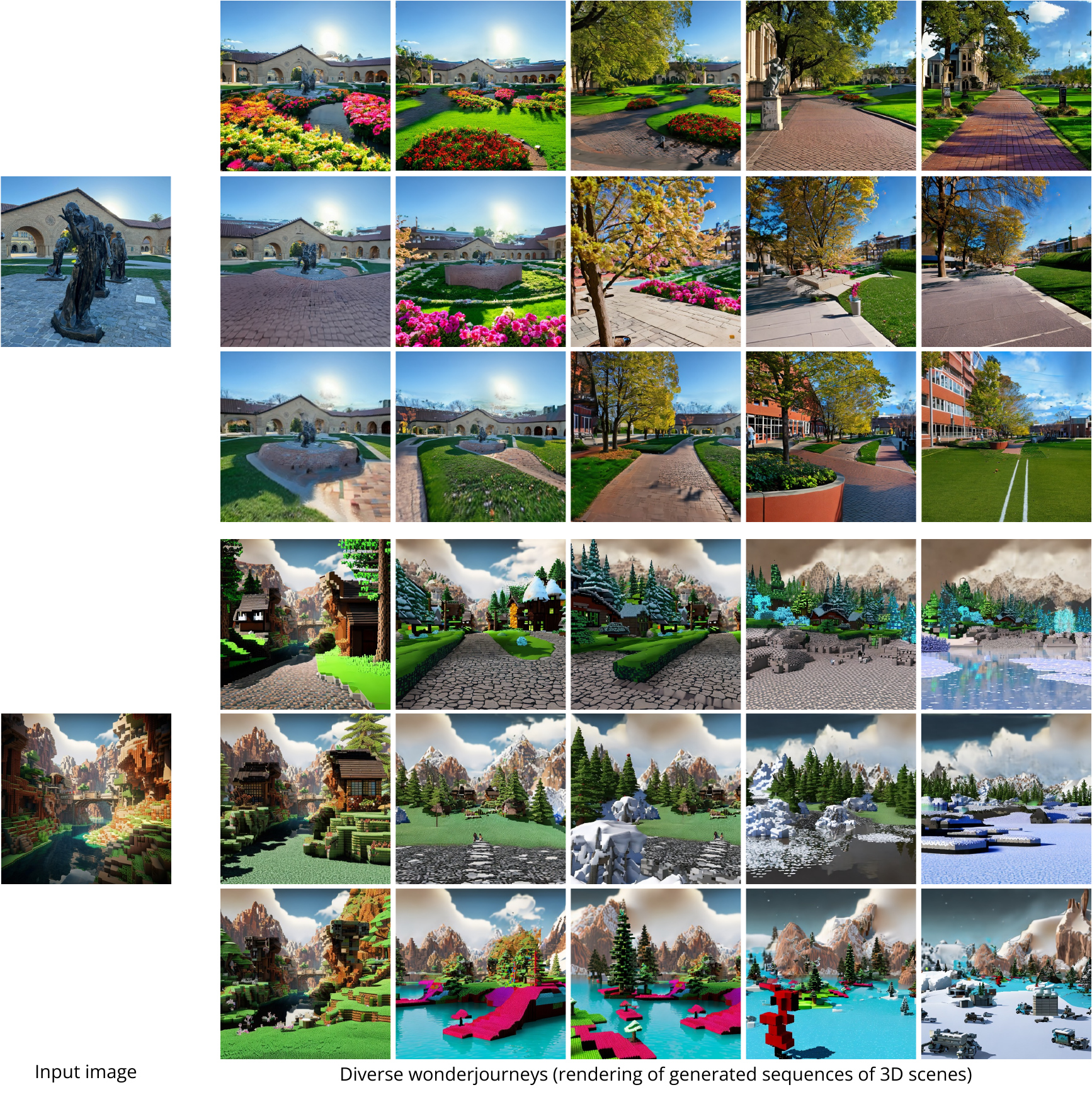}
\afterfig 
\caption{\textbf{Qualitative results} for diverse journeys generated from the same input image, showing that \model can go everywhere. The input in the top example is a real photo.}
\label{fig:diverse_results}
\vspace{-0.2cm}

\end{figure*}

\myparagraph{Depth refinement.} To enhance the depth discontinuity across object boundaries, we model scene elements with frontal planes when the elements have a limited disparity range. We use SAM~\citep{Kirillov_2023_ICCV} to generate pixel grouping segments $\{\texttt{seg}_j\}_{j=1}^{N_s}$ where $\texttt{seg}_j\in\{0,1\}^{H\times W}$ is a segment mask, sorted in descending order according to the size of the segment $\lVert \texttt{seg}_j \rVert$. We iteratively refine the estimated depth:
\begin{align}
\small
    \!\!\!\!\! \texttt{depth}[\texttt{seg}_j] \! \leftarrow  \!
    \begin{cases} 
\texttt{median}(\texttt{depth}[\texttt{seg}_j]), & \text{if } \Delta D_j \!<\! T, \\
\texttt{depth}[\texttt{seg}_j], & \text{otherwise},
    \end{cases}
\end{align}
for $j=1,\cdots, N_s$, where $\texttt{depth}\in \R_{+}^{H\times W}$ is initialized with the estimated monocular depth, $\texttt{median}(\cdot)$ is a function that returns the median value of the input set, $\Delta D_j=\texttt{max}(\texttt{disparity}[\texttt{seg}_j]) - \texttt{min}(\texttt{disparity}[\texttt{seg}_j])$ denotes the disparity (the reciprocal of \texttt{depth}) range within the segment $\texttt{seg}_j$. We keep the estimated depth of segments with high disparity range as they do not fit to a frontal-plane, such as roads. Note that the idea of frontal-plane modeling has also been explored in 3D Ken-Burns~\citep{niklaus20193d} with selected semantic classes such as car and people. 
In contrast, as we target at general scenes with diverse styles, we use the criterion of the disparity range for keeping estimated depth instead of selected semantic classes.

To handle the sky depth which is always underestimated, we use OneFormer~\citep{jain2023oneformer} to segment sky region and assign a high depth value to it. However, this results in inaccurate depth estimates along the sky boundary; if we were to naively use the output segmentation, these errors result in accumulated severe artifacts in later scenes. To resolve it, we simply remove points along the sky boundary. Besides, we find that depth at distant pixels are generally not reliable. Thus, we also set a far background plane with depth $F$ that cuts off all pixels' depth beyond it.

\myparagraph{Description-guided scene generation.}\label{sec:vision}
To generate a new scene that is connected to the current scene, we place a camera $C_{i+1}$ with an appropriate distance to the current camera $C_i$. 
As shown in \fig{\ref{fig:vision_part}}, we render the partial image $\hat{I}_{i+1}$ (more details on the camera and the renderer are in Appendix~\ref{sec:renderer}) and outpaint it with a text-guided outpainter to generate a new scene image $I_{i+1}$:
\aroundeqn
\begin{align}
    I_{i+1} = g_\text{outpaint}(\hat{I}_{i+1}, \mathcal{S}_{i+1}),
\end{align}
\aroundeqn
where we use the Stable Diffusion model~\citep{rombach2022high} for $g_\text{outpaint}$ in our experiments. 
Note, we purposefully place the new camera at a location that creates enough empty space in the rendered image. We empirically find that text-guided outpainters tend to avoid generating partial objects, likely due to their curated training image datasets, which tend to not include truncated or partial objects. Leaving too little empty space therefore results in just a simple extrapolation of the partial image $\hat{I}_{i+1}$ and a lack of adherence to the text prompt $\mathcal{S}_{i+1}$, especially in regards to new objects. After generating the new scene image, we lift it to 3D by estimating and refining depth for it, and we obtain the new point cloud $\hat{\mathcal{P}}_{i+1} = \mathcal{P}_i \cup \mathcal{P}'_{i+1}$ where $\mathcal{P}'_{i+1}$ denotes the additional points from unprojecting the outpainted pixels.

\myparagraph{New scene registration by depth consistency.} However, as the depth estimator is unaware of geometry constraints, the depth for points in $\mathcal{P}'$ generally do not align with $\mathcal{P}_i$. Thus, we adapt the depth estimator by a depth alignment loss:
\aroundeqn
\begin{align}
\footnotesize
    \Loss_\text{depth} = \max(0, \mathcal{D}^*_\text{bg}-\mathcal{D}'_\text{bg}) + \lVert\mathcal{D}^*_\text{fg}-\mathcal{D}'_\text{fg}\rVert,
\end{align}
\aroundeqn
where $\mathcal{D}^*_\text{bg}$ denotes the analytically computed depth of background pixels from $I_i$, $\mathcal{D}'_\text{bg}$ denotes the estimated depth for pixels corresponding to $\mathcal{D}^*_\text{bg}$, $\mathcal{D}^*_\text{fg}$ denotes the computed depth of all other visible pixels from $I_i$, and $\mathcal{D}'_\text{fg}$ denotes the estimated depth for pixels corresponding to $\mathcal{D}^*_\text{fg}$.

\myparagraph{Occlusion handling by re-rendering consistency.}
Another geometric inconsistency is that disocclusion regions can have a lower depth values than their occluders, as the depth estimator is not aware of this 3D geometric constraint. We highlight the wrongly estimated disocclusion depth in the refined depth in \fig{\ref{fig:vision_part}}. To resolve this issue, we re-render the new scene $\hat{\mathcal{P}}_{i+1}$ at the camera $C_i$ and detect all inconsistent pixels. At each inconsistent pixel, we move back all the rasterized additional points from $\mathcal{P}'_{i+1}$ that have lower depth values than the one point from $\mathcal{P}_i$. This removes the disocclusion inconsistency and guarantees that the disocclusion comes after the occluder.

\myparagraph{Scene completion.}
We obtain the final point cloud $\mathcal{P}_{i+1}$ by adding more points to $\hat{\mathcal{P}}_{i+1}$. We add points by repeating the following ``complete-as-you-go'' process: we place an additional camera along a camera trajectory connecting the new scene to the current scene, render a partial image at that camera, outpaint the image, and add the additional points to the point cloud.
Note that in our visual scene generation formulation in \eqn{\ref{eqn:visual}}, one can replace the image input $I_i$ with the point cloud $\mathcal{P}_i$ from the current scene, forming a persistent scene representation. This allows a trade-off between 3D persistence and empirical requirements. In practice, maintaining a large point cloud leads to prohibitively many points that require a large amount of GPU memory when generating a long trajectory of high-resolution scenes. Thus in our experiments we take the image formulation.

\begin{figure*}
    \centering
    \vspace{-0.8cm}
    \includegraphics[width=0.98\linewidth]{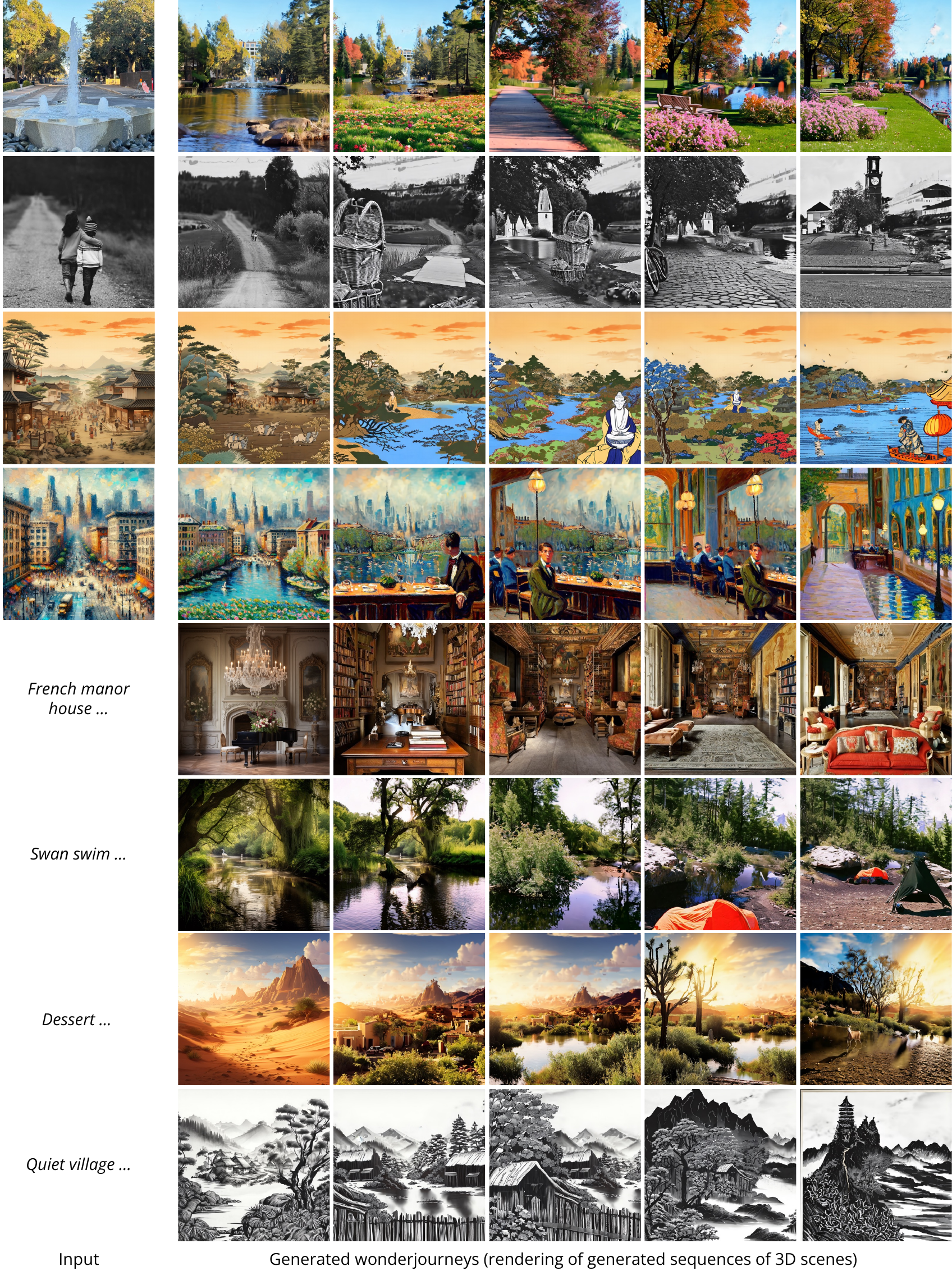}
    \vspace{-0.3cm}
    \caption{\textbf{From diverse starting scenes with different styles}, \model generates a sequence of diverse yet coherent 3D scenes, showing that it can go from anywhere to everywhere (\eg,~nature, village, city, indoor, or fantasy). \textbf{The inputs in top two rows are real photos. We strongly encourage the reader to see the video results in the project website.}}
    \label{fig:main_results}
    \afterfig
\end{figure*}

\subsection{Visual validation}
\aftersubsec

\ignore{Occasionally, we encounter undesirable geometry in the generated scene images, such as photo borders, frames, or out-of-focus blurry objects. Thus, we propose a validation step to identify and reject these undesired generated scenes.}

Empirically, in a large portion of generated photos and paintings, a painting frame or a photo border appears, disrupting the geometry consistency. Additionally, there are often unwanted blurry out-of-focus objects near the borders of the generated images. Thus, we propose a validation step to identify and reject these undesired generated scenes.

We formulate this as a text-based detection problem, where our objective is to detect a set of predefined undesirable effects in the generated scene image. We reject and regenerate the scene image if any unwanted effect is detected. Specifically, right after we generate a new scene image $I_{i+1}$, we immediately feed it to a VLM and prompt it with the query $\mathcal{J}^t_\text{detect}$ = \textit{``Is there any $X_t$ in this image?''} where $X_t\in \{ X_1, \cdots, X_T\}$ is an unwanted effect specified by text, such as ``photo border'', ``painting frame'', or ``out-of-focus objects''. If any unwanted effect is detected, we regenerate $I_{i+1}$ with a new description $\mathcal{S}_{i+1}$ or a new random seed.

\ignore{ \ky{we cannot address this; maybe leave it as a limitation.}
\myparagraph{Floaters from incomplete objects}
- Talk about how there are a few objects that can be generated incompletely. Namely trees. These objects need to be identified and removed, otherwise the depth warping can cause them to ``float'' in the next key-frame. }

\begin{figure*}
    \centering
    \includegraphics[width=0.95\textwidth]{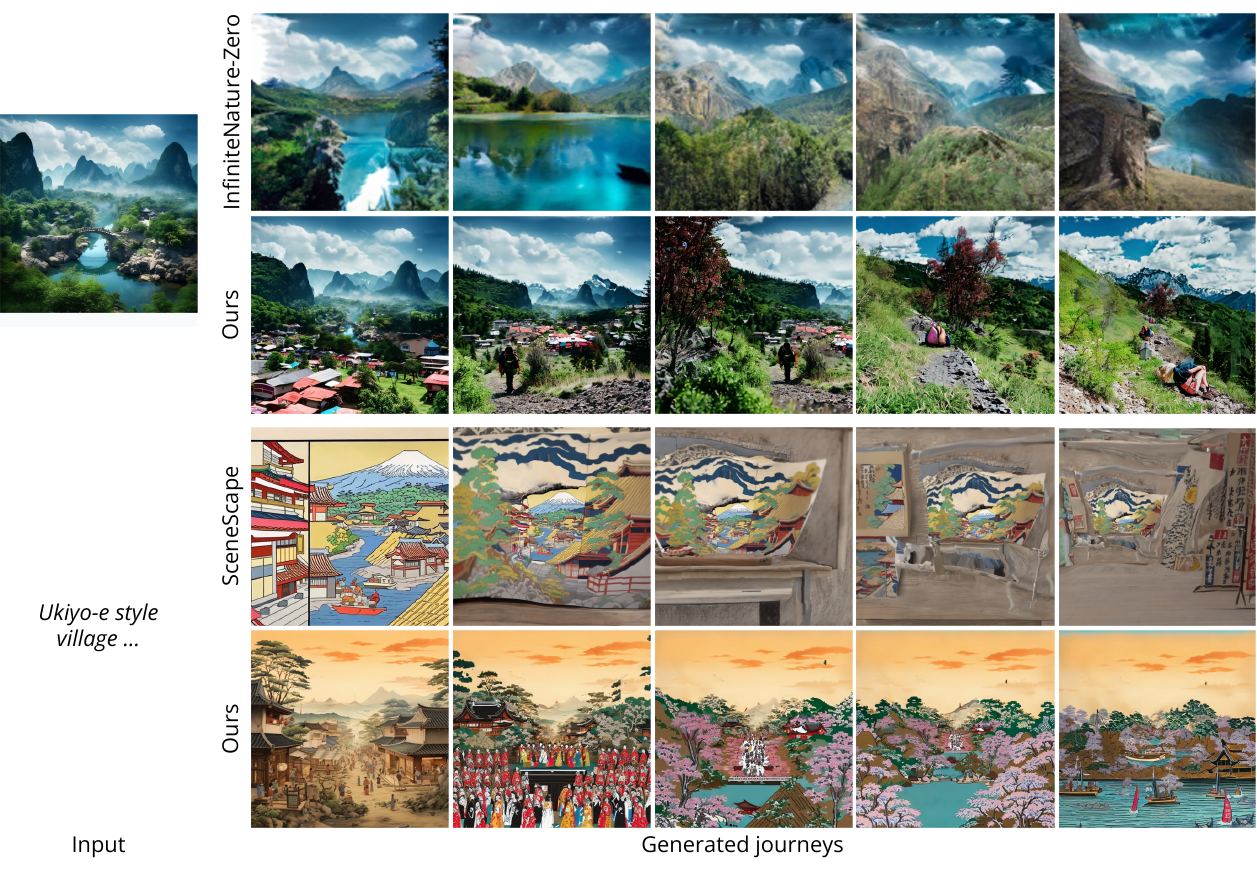}
    \vspace{-0.25cm}
    \caption{\textbf{Comparison with InfiniteNature-Zero~\citep{li2022infinitenature} and SceneScape~\citep{fridman2023scenescape}.} Note that InfiniteNature-Zero is trained on nature photos, so we only compare with it using photorealistic nature images as input.}
    \label{fig:baseline}
    \afterfig 
\end{figure*}
\section{Experiments}
\aftersec

\myparagraph{Dataset and baselines.}
Since the \problem is a new task without an existing dataset, we use a mixture of photos taken by ourselves, copyright-free photos from online, and generated examples, for evaluation in our experiments. We perform the pairing process by DALL-E 3~\citep{betker2023improving} for text-to-image pairing. 
We consider two state-of-the-art perpetual view generation methods as our baseline: image-based InfiniteNature-Zero and text-based SceneScape. %

\myparagraph{Qualitative demonstration.}
We show qualitative examples of the generated journey across different scenes and different styles in \fig{\ref{fig:teaser}} and \fig{\ref{fig:main_results}}. These results show that \model is able to generate diverse yet coherently connected scenes from various types of input images, i.e., it can go from anywhere. We show more examples in the Appendix. We further show examples of diverse generation samples from the same input in \fig{\ref{fig:diverse_results}}. These diverse generated journeys suggest that \model supports going to different destinations at each run.

\myparagraph{Additional evaluations.}
We show additional qualitative results in Appendix~\ref{sec:real_results}, longer ``wonderjourneys'' (up to $30$ scenes) in Appendix~\ref{sec:longer_journey}, controlled ``wonderjourneys'' (i.e., using user-provided full text, such as poems and haiku, instead of LLM-generated text guidance) in Appendix~\ref{sec:controlled}, and ablation studies in Appendix~\ref{sec:ablation}.

\begin{table}[t]
    \centering
    \footnotesize    
    \caption{Human preference of ours over baseline on diversity, visual quality, scene complexity, and  overall interesting-ness.}
    \aftertab
    \label{tbl:human}
    \begin{tabular}{l|cccc}
         & Div. & Qual. & Compl. & Overall  \\
         \midrule
      Ours over InfiniteNature-Zero  & 92.7\% & 94.9\% & 91.5\% & 88.6\%  \\
      Ours over SceneScape & 88.8\% & 83.4\% & 80.0\% & 90.3\%
    \end{tabular}
    \aftertab
\end{table}

\myparagraph{Human preference evaluation.}
Since a main application of \model is for creative and entertainment purposes, we focus on human preference evaluation as our quantitative metrics, using the following four axes: diversity of generated scenes in a single journey, visual quality, scene complexity, and overall interesting-ness.
We generate videos following each approach's own camera trajectories setup. 
Since InfiniteNature-Zero is trained on nature photos, we only compare to it using photorealistic nature images. 
For SceneScape, since it is text-based, we can use $3$ examples of different styles for comparison. 
We show a side-by-side comparison of videos generated by \model and a baseline with randomized positions. We then ask one binary-choice question at a time.
such as \textit{``Compare the two videos below. Which video has a higher \textbf{diversity}? That is, which video shows more various different places?''}.
We recruited $400$ participants, $200$ for the comparison with InfiniteNature-Zero and $200$ for SceneScape. Each participant answers $12$ questions. We provide more details in Appendix~\ref{sec:human_study}. 

As shown in \tbl{\ref{tbl:human}}, \model is strongly preferred over both baselines on all four axes. 
InfiniteNature-Zero synthesizes only nature scenes, as shown in \fig{\ref{fig:baseline}}, while \model generates more diverse scenes and objects such as mountaineers and houses that are naturally connected to the starting nature scene. SceneScape tends to generate cave-like scenes due to the usage of a textured mesh, and thus all examples converges to caves. Also, as discussed in Section~\ref{sec:vision}, SceneScape tends to not generate new objects due to limited white space. All these factors might contribute to the much greater user preference for \model.

\section{Conclusion}
\aftersec
\vspace{0.1cm}

We introduce \modelfull to generate a long sequence of diverse yet coherently connected 3D scenes starting at any user provided location.
\modelfull achieves compelling, diverse visual results across various scene types and different styles, enabling users to journey through their own adventures in the generated ``wonderjourneys''.

\vspace{-0.4cm}

\paragraph{Acknowledgments.} This work was supported by NSF RI \#2211258, ONR N00014-23-1-2355, and ONR YIP N00014-24-1-2117. The work was done in part when Hong-Xing Yu was a student researcher at Google and has been supported by gift funding and GCP credits from Google.

{
    \small
    \bibliographystyle{ieeenat_fullname}
    \bibliography{reference}

\begin{thebibliography}{43}
\providecommand{\natexlab}[1]{#1}
\providecommand{\url}[1]{\texttt{#1}}
\expandafter\ifx\csname urlstyle\endcsname\relax
  \providecommand{\doi}[1]{doi: #1}\else
  \providecommand{\doi}{doi: \begingroup \urlstyle{rm}\Url}\fi

\bibitem[Aleotti et~al.(2021)Aleotti, Tosi, Ramirez, Poggi, Salti, Mattoccia, and Di~Stefano]{aleotti2021neural}
Filippo Aleotti, Fabio Tosi, Pierluigi~Zama Ramirez, Matteo Poggi, Samuele Salti, Stefano Mattoccia, and Luigi Di~Stefano.
\newblock Neural disparity refinement for arbitrary resolution stereo.
\newblock In \emph{2021 International Conference on 3D Vision (3DV)}, pages 207--217. IEEE, 2021.

\bibitem[Bautista et~al.(2022)Bautista, Guo, Abnar, Talbott, Toshev, Chen, Dinh, Zhai, Goh, Ulbricht, et~al.]{bautista2022gaudi}
Miguel~Angel Bautista, Pengsheng Guo, Samira Abnar, Walter Talbott, Alexander Toshev, Zhuoyuan Chen, Laurent Dinh, Shuangfei Zhai, Hanlin Goh, Daniel Ulbricht, et~al.
\newblock Gaudi: A neural architect for immersive 3d scene generation.
\newblock \emph{Advances in Neural Information Processing Systems}, 35:\penalty0 25102--25116, 2022.

\bibitem[Betker et~al.(2023)Betker, Goh, Jing, Brooks, Wang, Li, Ouyang, Zhuang, Lee, Guo, Manassra, Dhariwal, Chu, Jiao, and Ramesh]{betker2023improving}
James Betker, Gabriel Goh, Li Jing, Tim Brooks, Jianfeng Wang, Linjie Li, Long Ouyang, Juntang Zhuang, Joyce Lee, Yufei Guo, Wesam Manassra, Prafulla Dhariwal, Casey Chu, Yunxin Jiao, and Aditya Ramesh.
\newblock Improving image generation with better captions.
\newblock \emph{Technical report}, 2023.

\bibitem[Cai et~al.(2023)Cai, Chan, Peng, Shahbazi, Obukhov, Van~Gool, and Wetzstein]{cai2022diffdreamer}
Shengqu Cai, Eric~Ryan Chan, Songyou Peng, Mohamad Shahbazi, Anton Obukhov, Luc Van~Gool, and Gordon Wetzstein.
\newblock {DiffDreamer}: {T}owards consistent unsupervised single-view scene extrapolation with conditional diffusion models.
\newblock In \emph{ICCV}, 2023.

\bibitem[Chai et~al.(2023)Chai, Tucker, Li, Isola, and Snavely]{chai2023persistent}
Lucy Chai, Richard Tucker, Zhengqi Li, Phillip Isola, and Noah Snavely.
\newblock Persistent nature: A generative model of unbounded 3d worlds.
\newblock In \emph{Proceedings of the IEEE/CVF Conference on Computer Vision and Pattern Recognition}, pages 20863--20874, 2023.

\bibitem[Chan et~al.(2021)Chan, Monteiro, Kellnhofer, Wu, and Wetzstein]{chan2021pi}
Eric~R Chan, Marco Monteiro, Petr Kellnhofer, Jiajun Wu, and Gordon Wetzstein.
\newblock pi-gan: Periodic implicit generative adversarial networks for 3d-aware image synthesis.
\newblock In \emph{Proceedings of the IEEE/CVF conference on computer vision and pattern recognition}, pages 5799--5809, 2021.

\bibitem[Chan et~al.(2022)Chan, Lin, Chan, Nagano, Pan, De~Mello, Gallo, Guibas, Tremblay, Khamis, et~al.]{chan2022efficient}
Eric~R Chan, Connor~Z Lin, Matthew~A Chan, Koki Nagano, Boxiao Pan, Shalini De~Mello, Orazio Gallo, Leonidas~J Guibas, Jonathan Tremblay, Sameh Khamis, et~al.
\newblock Efficient geometry-aware 3d generative adversarial networks.
\newblock In \emph{Proceedings of the IEEE/CVF Conference on Computer Vision and Pattern Recognition}, pages 16123--16133, 2022.

\bibitem[Chan et~al.(2023)Chan, Nagano, Chan, Bergman, Park, Levy, Aittala, De~Mello, Karras, and Wetzstein]{chan2023genvs}
Eric~R Chan, Koki Nagano, Matthew~A Chan, Alexander~W Bergman, Jeong~Joon Park, Axel Levy, Miika Aittala, Shalini De~Mello, Tero Karras, and Gordon Wetzstein.
\newblock Genvs: Generative novel view synthesis with 3d-aware diffusion models, 2023.

\bibitem[Cheng et~al.(2023)Cheng, Lee, Tulyakov, Schwing, and Gui]{cheng2023sdfusion}
Yen-Chi Cheng, Hsin-Ying Lee, Sergey Tulyakov, Alexander~G Schwing, and Liang-Yan Gui.
\newblock Sdfusion: Multimodal 3d shape completion, reconstruction, and generation.
\newblock In \emph{Proceedings of the IEEE/CVF Conference on Computer Vision and Pattern Recognition}, pages 4456--4465, 2023.

\bibitem[DeVries et~al.(2021)DeVries, Bautista, Srivastava, Taylor, and Susskind]{devries2021unconstrained}
Terrance DeVries, Miguel~Angel Bautista, Nitish Srivastava, Graham~W. Taylor, and Joshua~M. Susskind.
\newblock Unconstrained scene generation with locally conditioned radiance fields.
\newblock In \emph{ICCV}, 2021.

\bibitem[Fridman et~al.(2023)Fridman, Abecasis, Kasten, and Dekel]{fridman2023scenescape}
Rafail Fridman, Amit Abecasis, Yoni Kasten, and Tali Dekel.
\newblock Scenescape: Text-driven consistent scene generation.
\newblock \emph{arXiv preprint arXiv:2302.01133}, 2023.

\bibitem[Gong et~al.(2023)Gong, Pang, Cun, Xia, Chen, Wang, Zhang, Wang, Shan, and Yang]{gong2023talecrafter}
Yuan Gong, Youxin Pang, Xiaodong Cun, Menghan Xia, Haoxin Chen, Longyue Wang, Yong Zhang, Xintao Wang, Ying Shan, and Yujiu Yang.
\newblock Talecrafter: Interactive story visualization with multiple characters.
\newblock \emph{arXiv preprint arXiv:2305.18247}, 2023.

\bibitem[Gu et~al.(2021)Gu, Liu, Wang, and Theobalt]{gu2021stylenerf}
Jiatao Gu, Lingjie Liu, Peng Wang, and Christian Theobalt.
\newblock Stylenerf: A style-based 3d-aware generator for high-resolution image synthesis.
\newblock \emph{arXiv preprint arXiv:2110.08985}, 2021.

\bibitem[Hao et~al.(2021)Hao, Mallya, Belongie, and Liu]{hao2021gancraft}
Zekun Hao, Arun Mallya, Serge Belongie, and Ming-Yu Liu.
\newblock Gancraft: Unsupervised 3d neural rendering of minecraft worlds.
\newblock In \emph{Proceedings of the IEEE/CVF International Conference on Computer Vision}, pages 14072--14082, 2021.

\bibitem[H{\"o}llein et~al.(2023)H{\"o}llein, Cao, Owens, Johnson, and Nie{\ss}ner]{hollein2023text2room}
Lukas H{\"o}llein, Ang Cao, Andrew Owens, Justin Johnson, and Matthias Nie{\ss}ner.
\newblock Text2room: Extracting textured 3d meshes from 2d text-to-image models.
\newblock \emph{arXiv preprint arXiv:2303.11989}, 2023.

\bibitem[Huang et~al.(2023)Huang, Feng, Shi, Xu, Yu, and Yang]{huang2023free}
Hanzhuo Huang, Yufan Feng, Cheng Shi, Lan Xu, Jingyi Yu, and Sibei Yang.
\newblock Free-bloom: Zero-shot text-to-video generator with llm director and ldm animator.
\newblock \emph{arXiv preprint arXiv:2309.14494}, 2023.

\bibitem[Jain et~al.(2023)Jain, Li, Chiu, Hassani, Orlov, and Shi]{jain2023oneformer}
Jitesh Jain, Jiachen Li, MangTik Chiu, Ali Hassani, Nikita Orlov, and Humphrey Shi.
\newblock {OneFormer: One Transformer to Rule Universal Image Segmentation}.
\newblock In \emph{CVPR}, 2023.

\bibitem[Kaneva et~al.(2010)Kaneva, Sivic, Torralba, Avidan, and Freeman]{kaneva2010infinite}
Biliana Kaneva, Josef Sivic, Antonio Torralba, Shai Avidan, and William~T Freeman.
\newblock Infinite images: Creating and exploring a large photorealistic virtual space.
\newblock \emph{Proceedings of the IEEE}, 98\penalty0 (8):\penalty0 1391--1407, 2010.

\bibitem[Kirillov et~al.(2023)Kirillov, Mintun, Ravi, Mao, Rolland, Gustafson, Xiao, Whitehead, Berg, Lo, Dollar, and Girshick]{Kirillov_2023_ICCV}
Alexander Kirillov, Eric Mintun, Nikhila Ravi, Hanzi Mao, Chloe Rolland, Laura Gustafson, Tete Xiao, Spencer Whitehead, Alexander~C. Berg, Wan-Yen Lo, Piotr Dollar, and Ross Girshick.
\newblock Segment anything.
\newblock In \emph{ICCV}, pages 4015--4026, 2023.

\bibitem[Li et~al.(2022{\natexlab{a}})Li, Zheng, Wang, Li, Zheng, and Tao]{li20223ddesigner}
Gang Li, Heliang Zheng, Chaoyue Wang, Chang Li, Changwen Zheng, and Dacheng Tao.
\newblock 3ddesigner: Towards photorealistic 3d object generation and editing with text-guided diffusion models.
\newblock \emph{arXiv preprint arXiv:2211.14108}, 2022{\natexlab{a}}.

\bibitem[Li et~al.(2022{\natexlab{b}})Li, Wang, Snavely, and Kanazawa]{li2022infinitenature}
Zhengqi Li, Qianqian Wang, Noah Snavely, and Angjoo Kanazawa.
\newblock Infinitenature-zero: Learning perpetual view generation of natural scenes from single images.
\newblock In \emph{European Conference on Computer Vision}, pages 515--534. Springer, 2022{\natexlab{b}}.

\bibitem[Lin et~al.(2023{\natexlab{a}})Lin, Gao, Tang, Takikawa, Zeng, Huang, Kreis, Fidler, Liu, and Lin]{lin2023magic3d}
Chen-Hsuan Lin, Jun Gao, Luming Tang, Towaki Takikawa, Xiaohui Zeng, Xun Huang, Karsten Kreis, Sanja Fidler, Ming-Yu Liu, and Tsung-Yi Lin.
\newblock Magic3d: High-resolution text-to-3d content creation.
\newblock In \emph{Proceedings of the IEEE/CVF Conference on Computer Vision and Pattern Recognition}, pages 300--309, 2023{\natexlab{a}}.

\bibitem[Lin et~al.(2023{\natexlab{b}})Lin, Zala, Cho, and Bansal]{lin2023videodirectorgpt}
Han Lin, Abhay Zala, Jaemin Cho, and Mohit Bansal.
\newblock Videodirectorgpt: Consistent multi-scene video generation via llm-guided planning.
\newblock \emph{arXiv preprint arXiv:2309.15091}, 2023{\natexlab{b}}.

\bibitem[Liu et~al.(2021)Liu, Tucker, Jampani, Makadia, Snavely, and Kanazawa]{liu2021infinite}
Andrew Liu, Richard Tucker, Varun Jampani, Ameesh Makadia, Noah Snavely, and Angjoo Kanazawa.
\newblock Infinite nature: Perpetual view generation of natural scenes from a single image.
\newblock In \emph{Proceedings of the IEEE/CVF International Conference on Computer Vision}, pages 14458--14467, 2021.

\bibitem[Liu et~al.(2023{\natexlab{a}})Liu, Wu, Zhong, Zhang, and Xie]{liu2023intelligent}
Chang Liu, Haoning Wu, Yujie Zhong, Xiaoyun Zhang, and Weidi Xie.
\newblock Intelligent grimm--open-ended visual storytelling via latent diffusion models.
\newblock \emph{arXiv preprint arXiv:2306.00973}, 2023{\natexlab{a}}.

\bibitem[Liu et~al.(2023{\natexlab{b}})Liu, Wu, Van~Hoorick, Tokmakov, Zakharov, and Vondrick]{liu2023zero}
Ruoshi Liu, Rundi Wu, Basile Van~Hoorick, Pavel Tokmakov, Sergey Zakharov, and Carl Vondrick.
\newblock Zero-1-to-3: Zero-shot one image to 3d object.
\newblock In \emph{Proceedings of the IEEE/CVF International Conference on Computer Vision}, pages 9298--9309, 2023{\natexlab{b}}.

\bibitem[Melas-Kyriazi et~al.(2023)Melas-Kyriazi, Laina, Rupprecht, and Vedaldi]{melas2023realfusion}
Luke Melas-Kyriazi, Iro Laina, Christian Rupprecht, and Andrea Vedaldi.
\newblock Realfusion: 360deg reconstruction of any object from a single image.
\newblock In \emph{Proceedings of the IEEE/CVF Conference on Computer Vision and Pattern Recognition}, pages 8446--8455, 2023.

\bibitem[Miangoleh et~al.(2021)Miangoleh, Dille, Mai, Paris, and Aksoy]{miangoleh2021boosting}
S~Mahdi~H Miangoleh, Sebastian Dille, Long Mai, Sylvain Paris, and Yagiz Aksoy.
\newblock Boosting monocular depth estimation models to high-resolution via content-adaptive multi-resolution merging.
\newblock In \emph{Proceedings of the IEEE/CVF Conference on Computer Vision and Pattern Recognition}, pages 9685--9694, 2021.

\bibitem[Nguyen-Phuoc et~al.(2019)Nguyen-Phuoc, Li, Theis, Richardt, and Yang]{nguyen2019hologan}
Thu Nguyen-Phuoc, Chuan Li, Lucas Theis, Christian Richardt, and Yong-Liang Yang.
\newblock Hologan: Unsupervised learning of 3d representations from natural images.
\newblock In \emph{Proceedings of the IEEE/CVF International Conference on Computer Vision}, pages 7588--7597, 2019.

\bibitem[Niemeyer and Geiger(2021)]{niemeyer2021giraffe}
Michael Niemeyer and Andreas Geiger.
\newblock Giraffe: Representing scenes as compositional generative neural feature fields.
\newblock In \emph{Proceedings of the IEEE/CVF Conference on Computer Vision and Pattern Recognition}, pages 11453--11464, 2021.

\bibitem[Niklaus et~al.(2019)Niklaus, Mai, Yang, and Liu]{niklaus20193d}
Simon Niklaus, Long Mai, Jimei Yang, and Feng Liu.
\newblock 3d ken burns effect from a single image.
\newblock \emph{ACM Transactions on Graphics (ToG)}, 38\penalty0 (6):\penalty0 1--15, 2019.

\bibitem[Or-El et~al.(2022)Or-El, Luo, Shan, Shechtman, Park, and Kemelmacher-Shlizerman]{or2022stylesdf}
Roy Or-El, Xuan Luo, Mengyi Shan, Eli Shechtman, Jeong~Joon Park, and Ira Kemelmacher-Shlizerman.
\newblock Stylesdf: High-resolution 3d-consistent image and geometry generation.
\newblock In \emph{Proceedings of the IEEE/CVF Conference on Computer Vision and Pattern Recognition}, pages 13503--13513, 2022.

\bibitem[Poole et~al.(2022)Poole, Jain, Barron, and Mildenhall]{poole2022dreamfusion}
Ben Poole, Ajay Jain, Jonathan~T Barron, and Ben Mildenhall.
\newblock Dreamfusion: Text-to-3d using 2d diffusion.
\newblock \emph{arXiv preprint arXiv:2209.14988}, 2022.

\bibitem[Raj et~al.(2023)Raj, Kaza, Poole, Niemeyer, Ruiz, Mildenhall, Zada, Aberman, Rubinstein, Barron, et~al.]{raj2023dreambooth3d}
Amit Raj, Srinivas Kaza, Ben Poole, Michael Niemeyer, Nataniel Ruiz, Ben Mildenhall, Shiran Zada, Kfir Aberman, Michael Rubinstein, Jonathan Barron, et~al.
\newblock Dreambooth3d: Subject-driven text-to-3d generation.
\newblock \emph{arXiv preprint arXiv:2303.13508}, 2023.

\bibitem[Ranftl et~al.(2022)Ranftl, Lasinger, Hafner, Schindler, and Koltun]{Ranftl2022}
Ren\'{e} Ranftl, Katrin Lasinger, David Hafner, Konrad Schindler, and Vladlen Koltun.
\newblock Towards robust monocular depth estimation: Mixing datasets for zero-shot cross-dataset transfer.
\newblock \emph{IEEE Transactions on Pattern Analysis and Machine Intelligence}, 44\penalty0 (3), 2022.

\bibitem[Rombach et~al.(2022)Rombach, Blattmann, Lorenz, Esser, and Ommer]{rombach2022high}
Robin Rombach, Andreas Blattmann, Dominik Lorenz, Patrick Esser, and Bj{\"o}rn Ommer.
\newblock High-resolution diffusion models.
\newblock In \emph{Proceedings of the IEEE/CVF conference on computer vision and pattern recognition}, pages 10684--10695, 2022.

\bibitem[Sargent et~al.(2023{\natexlab{a}})Sargent, Koh, Zhang, Chang, Herrmann, Srinivasan, Wu, and Sun]{sargent2023vq3d}
Kyle Sargent, Jing~Yu Koh, Han Zhang, Huiwen Chang, Charles Herrmann, Pratul Srinivasan, Jiajun Wu, and Deqing Sun.
\newblock Vq3d: Learning a 3d-aware generative model on imagenet.
\newblock \emph{arXiv preprint arXiv:2302.06833}, 2023{\natexlab{a}}.

\bibitem[Sargent et~al.(2023{\natexlab{b}})Sargent, Li, Shah, Herrmann, Yu, Zhang, Chan, Lagun, Fei-Fei, Sun, et~al.]{sargent2023zeronvs}
Kyle Sargent, Zizhang Li, Tanmay Shah, Charles Herrmann, Hong-Xing Yu, Yunzhi Zhang, Eric~Ryan Chan, Dmitry Lagun, Li Fei-Fei, Deqing Sun, et~al.
\newblock Zeronvs: Zero-shot 360-degree view synthesis from a single real image.
\newblock \emph{arXiv preprint arXiv:2310.17994}, 2023{\natexlab{b}}.

\bibitem[Schwarz et~al.(2022)Schwarz, Sauer, Niemeyer, Liao, and Geiger]{schwarz2022voxgraf}
Katja Schwarz, Axel Sauer, Michael Niemeyer, Yiyi Liao, and Andreas Geiger.
\newblock Voxgraf: Fast 3d-aware image synthesis with sparse voxel grids.
\newblock \emph{Advances in Neural Information Processing Systems}, 35:\penalty0 33999--34011, 2022.

\bibitem[Tewari et~al.(2023)Tewari, Yin, Cazenavette, Rezchikov, Tenenbaum, Durand, Freeman, and Sitzmann]{tewari2023diffusion}
Ayush Tewari, Tianwei Yin, George Cazenavette, Semon Rezchikov, Joshua~B Tenenbaum, Fr{\'e}do Durand, William~T Freeman, and Vincent Sitzmann.
\newblock Diffusion with forward models: Solving stochastic inverse problems without direct supervision.
\newblock \emph{arXiv preprint arXiv:2306.11719}, 2023.

\bibitem[Tosi et~al.(2021)Tosi, Liao, Schmitt, and Geiger]{tosi2021smd}
Fabio Tosi, Yiyi Liao, Carolin Schmitt, and Andreas Geiger.
\newblock Smd-nets: Stereo mixture density networks.
\newblock In \emph{Proceedings of the IEEE/CVF Conference on Computer Vision and Pattern Recognition}, pages 8942--8952, 2021.

\bibitem[Wang et~al.(2021)Wang, Wu, Xie, Chen, and Prisacariu]{wang2021nerf}
Zirui Wang, Shangzhe Wu, Weidi Xie, Min Chen, and Victor~Adrian Prisacariu.
\newblock Nerf--: Neural radiance fields without known camera parameters.
\newblock \emph{arXiv preprint arXiv:2102.07064}, 2021.

\bibitem[Yu et~al.(2023)Yu, Forghani, Derpanis, and Brubaker]{yu2023long}
Jason~J Yu, Fereshteh Forghani, Konstantinos~G Derpanis, and Marcus~A Brubaker.
\newblock Long-term photometric consistent novel view synthesis with diffusion models.
\newblock \emph{arXiv preprint arXiv:2304.10700}, 2023.

\end{thebibliography}
}

\clearpage
\setcounter{page}{1}
\maketitleappendix

\appendix
\section{Overview}
We compile a set of video results in our project website. \textbf{We strongly encourage the reader to see these video results}. Please use a modern browser such as Chrome, since we use advanced JavaScript libraries to control the carousels and video auto-play.

In the following, we summarize the contents in this supplementary document:
\vspace{0.1cm}
\begin{itemize}
\setlength{\itemsep}{0.2em}
    \item $[$\sect{\ref{sec:real_results}}$]$ Additional qualitative results.
    \item $[$\sect{\ref{sec:longer_journey}}$]$ Longer ``wonderjourneys''. Each ``wonderjourney'' consists of $30$ generated scenes.
    \item $[$\sect{\ref{sec:controlled}}$]$ Controlled ``wonderjourneys'' using user-provided descriptions (rather than LLM-generated descriptions), such as poems, story abstracts, and haiku.
    \item $[$\sect{\ref{sec:renderer}}$]$ Additional details on our renderer, camera paths, and depth processing.
    \item $[$\sect{\ref{sec:ablation}}$]$ Ablation study on white space ratio, the visual validation, and depth processing.
    \item $[$\sect{\ref{sec:prompts}}$]$ Additional details on the LLM and VLM we use.
    \item $[$\sect{\ref{sec:human_study}}$]$ Details on human preference evaluation setting.
\end{itemize}

\section{Additional Results}
\label{sec:real_results}

\begin{figure*}
\includegraphics[width=\textwidth]{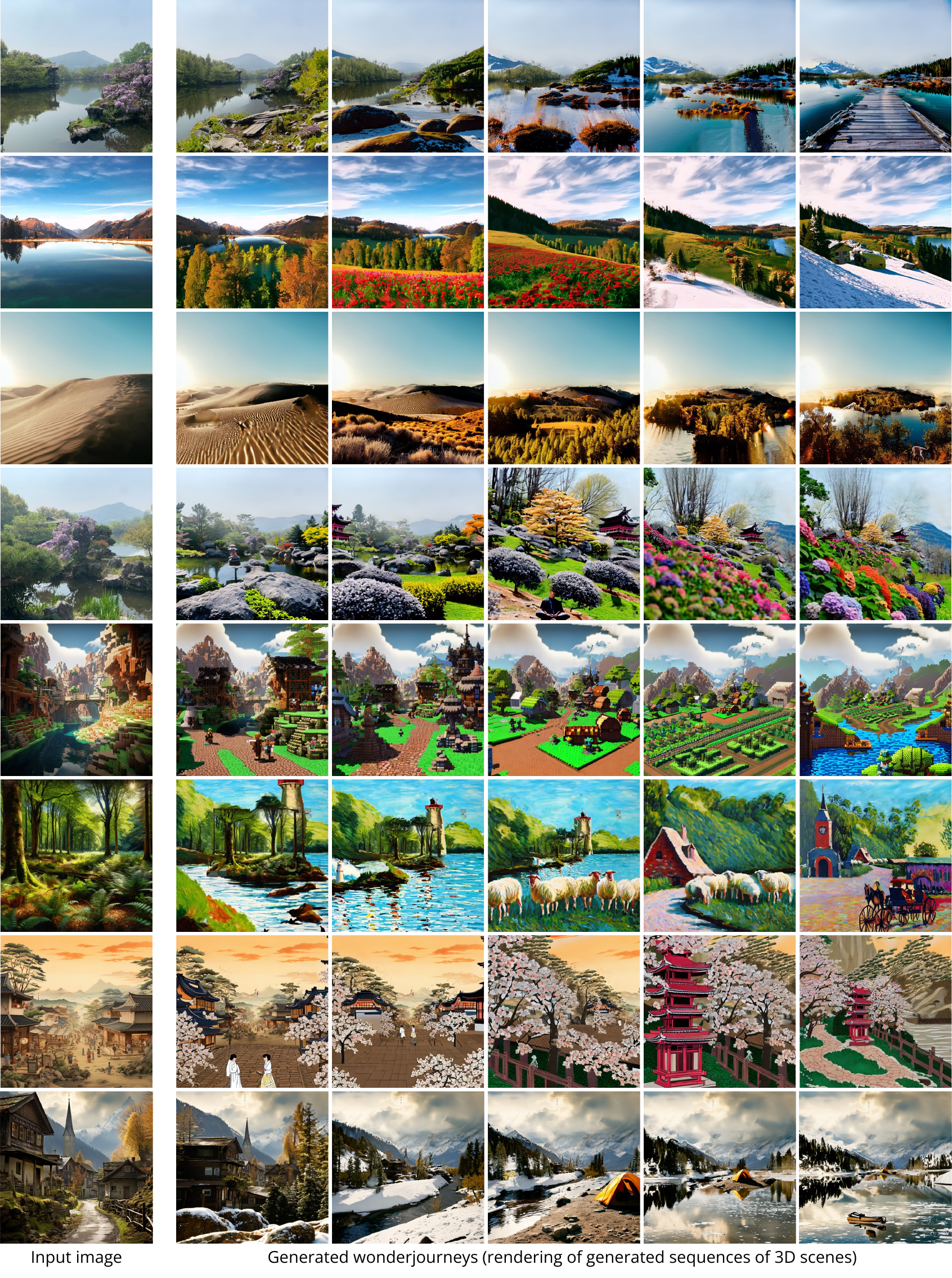}
\vspace{-0.3cm}
\afterfig 
\caption{\textbf{Additional qualitative results} for ``wonderjourneys''. The inputs in the top four rows are real photos. The inputs in top four rows are real photos.}
\label{fig:real_anywhere}
\end{figure*}
\begin{figure*}
\includegraphics[width=\textwidth]{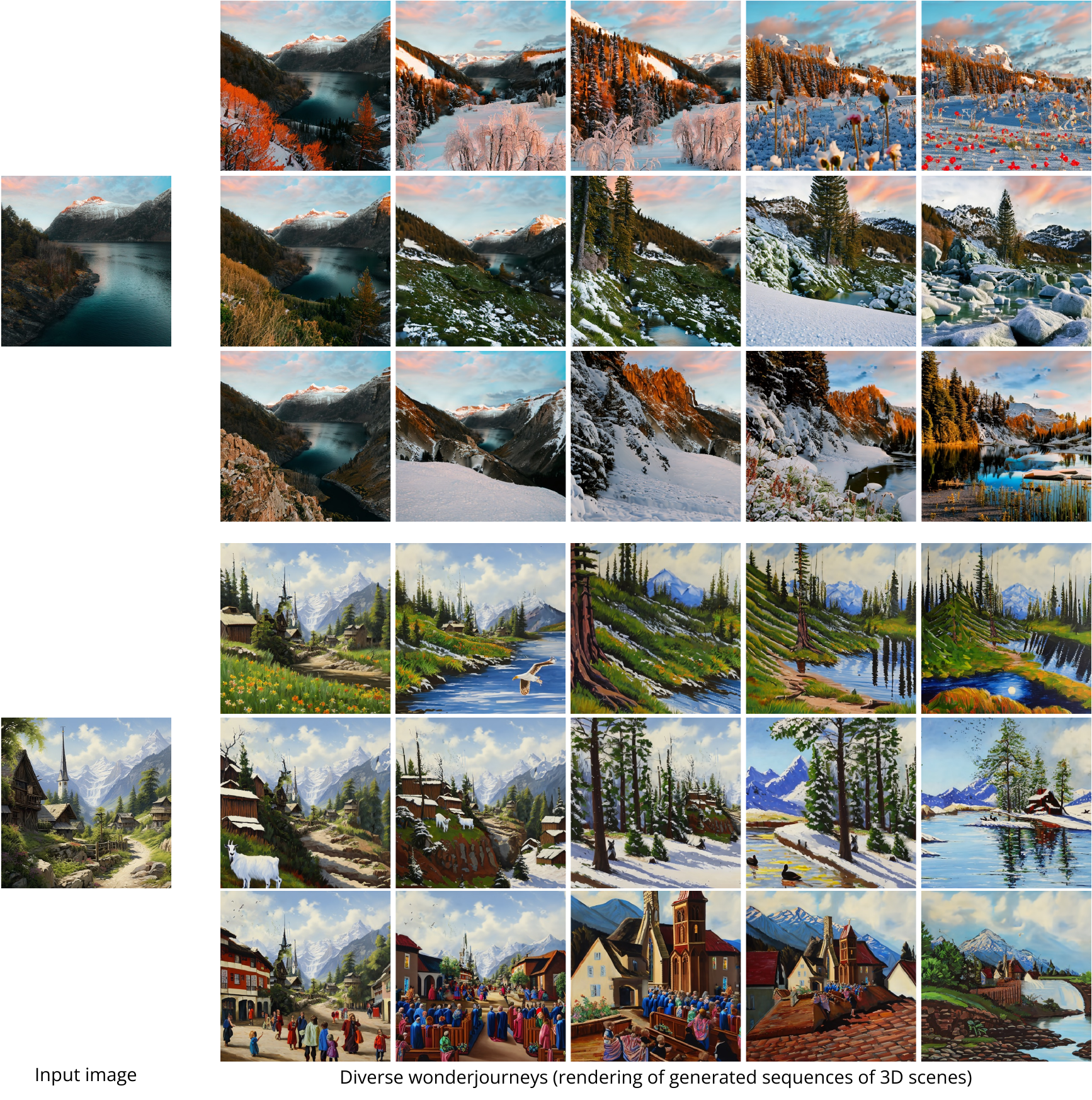}
\vspace{-0.3cm}
\afterfig 
\caption{\textbf{Qualitative results for diverse} ``wonderjourneys'' generated from the same input image. The inputs in the top example is a real photo.}
\label{fig:real_diverse}

\end{figure*}
We show additional results in 
\fig{\ref{fig:real_anywhere}} (going from anywhere) 
and \fig{\ref{fig:real_diverse}} (going to everywhere). 

\section{Longer ``Wonderjourneys''}\label{sec:longer_journey}

\begin{figure*}
\includegraphics[width=\textwidth]{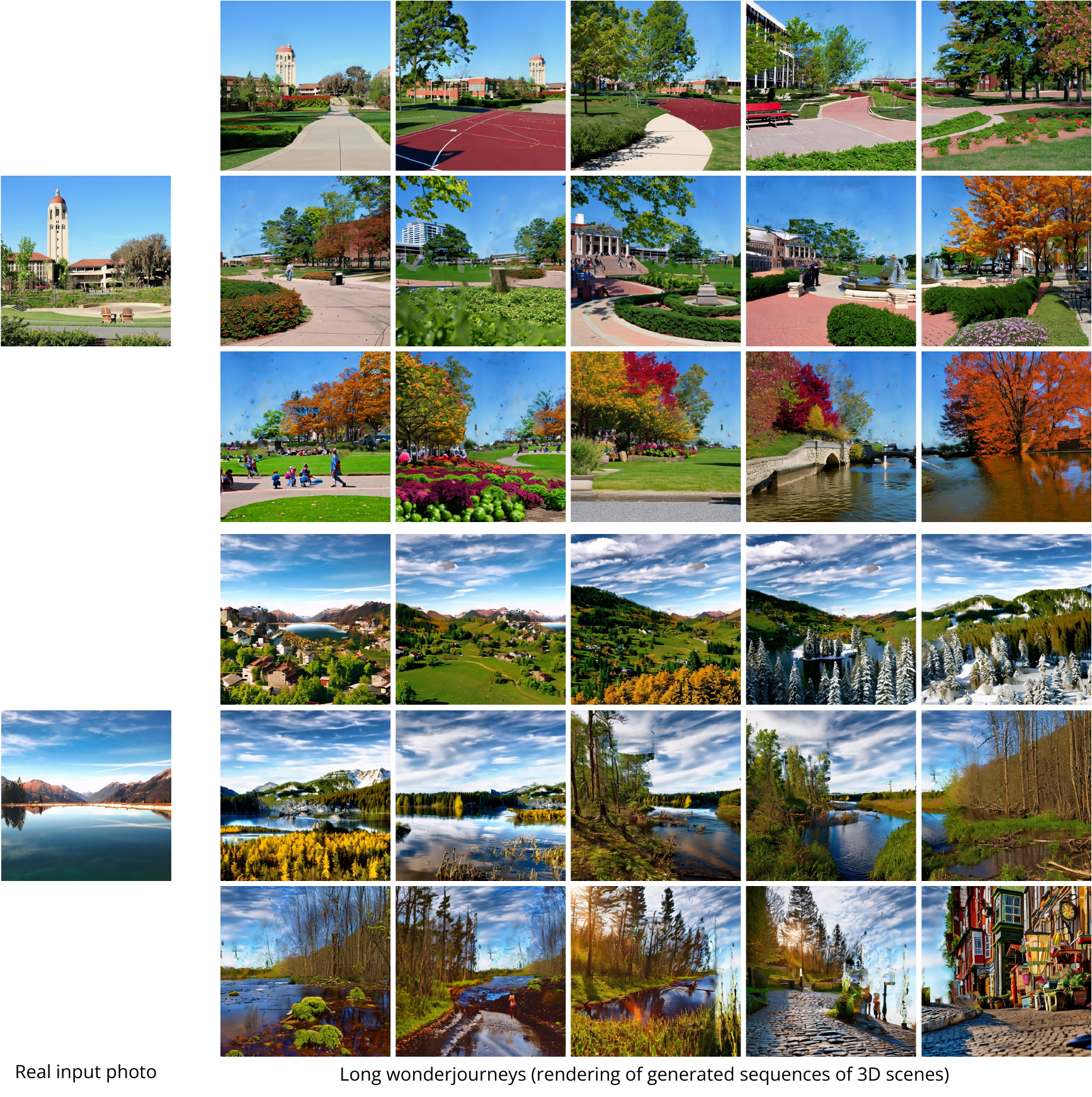}
\vspace{-0.3cm}
\afterfig 
\caption{\textbf{Qualitative results for long ``wonderjourneys''} generated from real input photos. Each long ``wonderjourney'' here consists of $30$ scenes, yet we show $15$ of them. See our website for video results.}
\label{fig:long_journey}
\end{figure*}
We show examples of longer ``wonderjourneys'' in \fig{\ref{fig:long_journey}}. We observe that the longer ``wonderjourneys'' allow including more diverse scenes with high visual quality.
\section{Controlled ``Wonderjourneys''}\label{sec:controlled}

\begin{figure*}
\includegraphics[width=\textwidth]{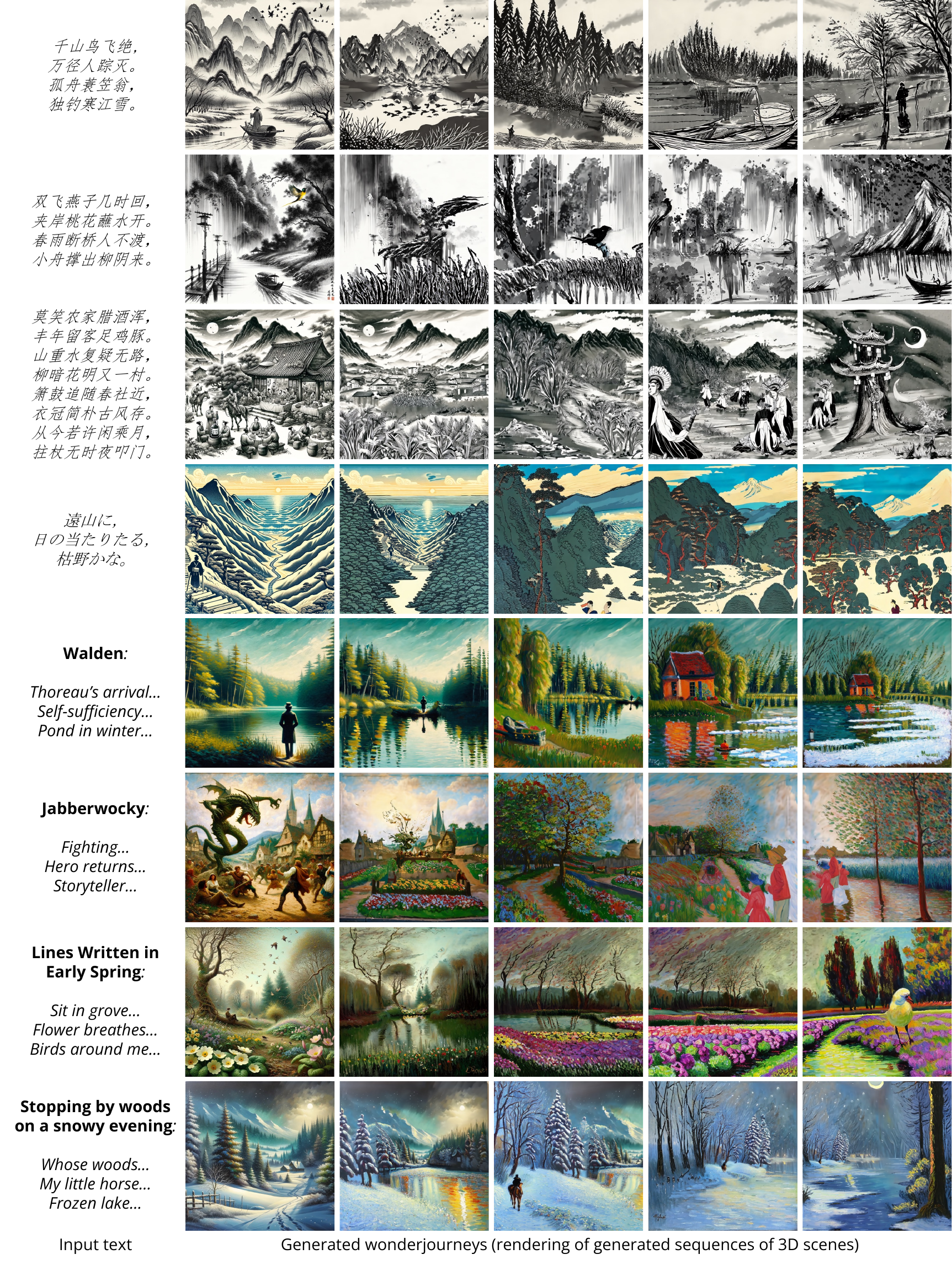}
\vspace{-0.3cm}
\afterfig 
\caption{\textbf{Qualitative results for controlled ``wonderjourneys''} generated from text descriptions (except for the ``Jabberwocky'' example where we manually pair it with an image). Each controlled ``wonderjourney'' here consists of $2N$ scenes where $N$ is the number of parts of the text (e.g., a classical Chinese poem often has $4$ or $8$ parts). See our website for video results.}
\label{fig:controlled}
\end{figure*}
We may replace the LLM-generated scene descriptions with user-provided descriptions to control the generated ``wonderjourneys''. For example, one can use poems, haiku, or story abstracts. We show examples of classical Chinese poems, haiku, a nonsense poem ``Jabberwocky'' from \emph{Alice's Adventures in Wonderland}, abstract of \emph{Walden} by Henry David Thoreau, \emph{Stopping by Woods on a Snowy Evening} by Robert Frost, and \emph{Lines Written in Early Spring} by William Wordsworth in \fig{\ref{fig:controlled}}.
\section{Details on Visual Scene Generation}\label{sec:renderer}
\myparagraph{Depth processing.} As mentioned in the main paper, we find that depths of sky and distant pixels are estimated incorrectly. This is a general issue in monocular depth estimators, although we choose to use MiDaS v3.1~\citep{Ranftl2022}. For the segmented sky pixels, we set the depth to $0.025$. For distant pixels, we set the background far plane $F=0.0015$. Since MiDaS is shift-invariant, we manually add a depth shift $0.0001$ to ensure that the near objects do not collapse to the optical center due to extremely small depth values. Note that all these values (and all the depth estimator-related values below) are specific to MiDaS v3.1, and it may need to be changed for other depth estimators due to different normalization schemes used in their respective training procedures.

The disparity threshold $T$ in depth refinement is set to $2$. Empirically, we use the $30\%$ and $70\%$ depth values instead of $\min$ and $\max$ depth values within a segment to compute $\Delta D_j$ to improve robustness due to segmentation inaccuracy. 

In new scene registration by depth consistency, we adapt MiDaS v3.1 for $200$ iterations with learning rate $0.000001$. In occlusion handling, we set the depth of disoccluders to $0.05$. In scene completion, we set the depth of newly inpainted points to be the same as the depth of its valid nearest neighbor pixel. 

\myparagraph{Sky processing.} Empirically, we find that sky segmentation is generally not accurate enough especially along the boundary of sharp shapes (such as tower spires) and complex shapes (such as tree leaves). Therefore, we use the following process to combat the inaccuracy. After using SAM to refine depth, we use an aggressively eroded sky segmentation to set depth values for sky pixels. Since SAM also segments sky slightly more accurately (although it often gives over-segments and it does not have semantic labels), we then use SAM to refine depth again to take advantage of the added accuracy. However, even SAM has difficulty in accurately segmenting complex sky-object boundaries. Thus, we dilate our outpainting mask a bit in the upper part of the image to cover the potentially inaccurately segmented boundary.

\myparagraph{Rendering.} We use a perspective pinhole camera model. To render the point cloud, we transform the points to a normalized device coordinate space and rasterize them to determine visibility. For each camera ray, we allow up to $8$ points to reside in the $z$-buffer, and composite them using the following softmax-based function to avoid alias:
\begin{align}
    \texttt{disparity} = \frac{\sum_i\exp(\texttt{disparity}_i)*\texttt{color}_i}{\sum_i\exp(\texttt{disparity}_i)},
\end{align}
where $i=1,\cdots,8$ indexes the points in the $z$-buffer. Higher disparity values put higher weights on nearer points. At the boundary of objects, occluded points can also provide some blending for anti-aliasing. Our image resolution is $512\times 512$.

\myparagraph{Camera paths.}
To generate visual scenes, we move our camera either following a straight line, or to do a rotation. For the straight line, we use a camera movement of $0.0005$ to the backward. For rotation, we move our camera with a rotation of $0.45$ radians with a translation of $0.0001$.

To generate the additional camera paths, we use the following rules: For a generated scene by rotation, we interpolate linearly among the camera rotation radians. For a generated scene by straight line, we linearly interpolate the translation. The additional cameras are also used in making our video results. Therefore, we also add a random sine perturbation to the height of the additional cameras. For the starting scene, the ending scene, and scenes right before and right after rotation happens, we add a linear acceleration process to make the video smoother.
\section{Ablation Study}\label{sec:ablation}
\begin{figure}
    \centering
    \includegraphics[width=\linewidth]{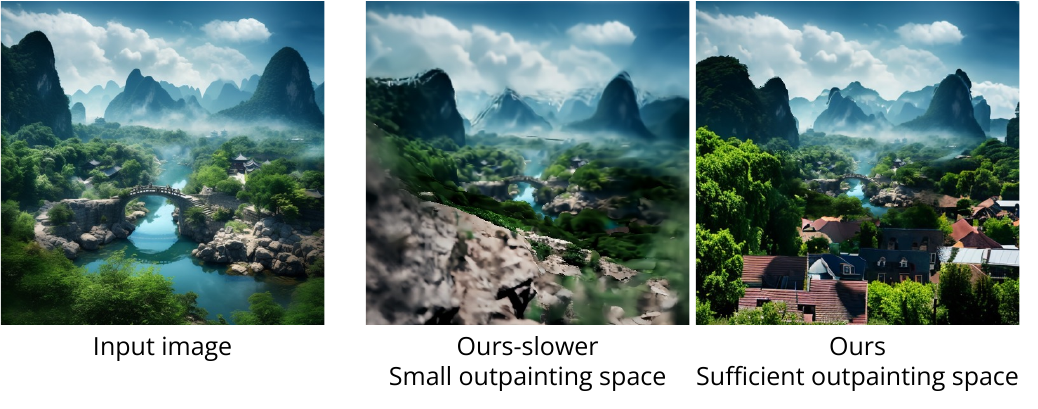}
    \vspace{-0.4cm}
    \caption{\textbf{Ablation comparison on sufficient outpainting space.} Both examples use the same scene descriptions including ``houses'' in it. Only ``Ours'' that has sufficient outpainting space can generate houses.}
    \label{fig:ablation_white_space}
\end{figure}
\myparagraph{Outpainting white space.} In \fig{\ref{fig:ablation_white_space}}, we show a comparison with the following variant, ``Ours-slower''. ``Ours-slower'' uses a $1/10$ speed and generating $10$ scenes following a straight line camera path (see \sect{\ref{sec:renderer}} for more details on the camera path), so that it ends up at the same camera location as ``Ours''. We use the same input image and the same scene description which includes ``houses'' in it. From \fig{\ref{fig:ablation_white_space}} we observe that when we have insufficient empty space for outpainting, new objects like houses do not appear. This empirical observation may be due to that in the curated training set of the Stable Diffusion model, there may be few partial cropped objects around the image borders. Thus, the outpainting model does not prefer generating partial objects along the borders.

\begin{figure}
    \centering
    \includegraphics[width=\linewidth]{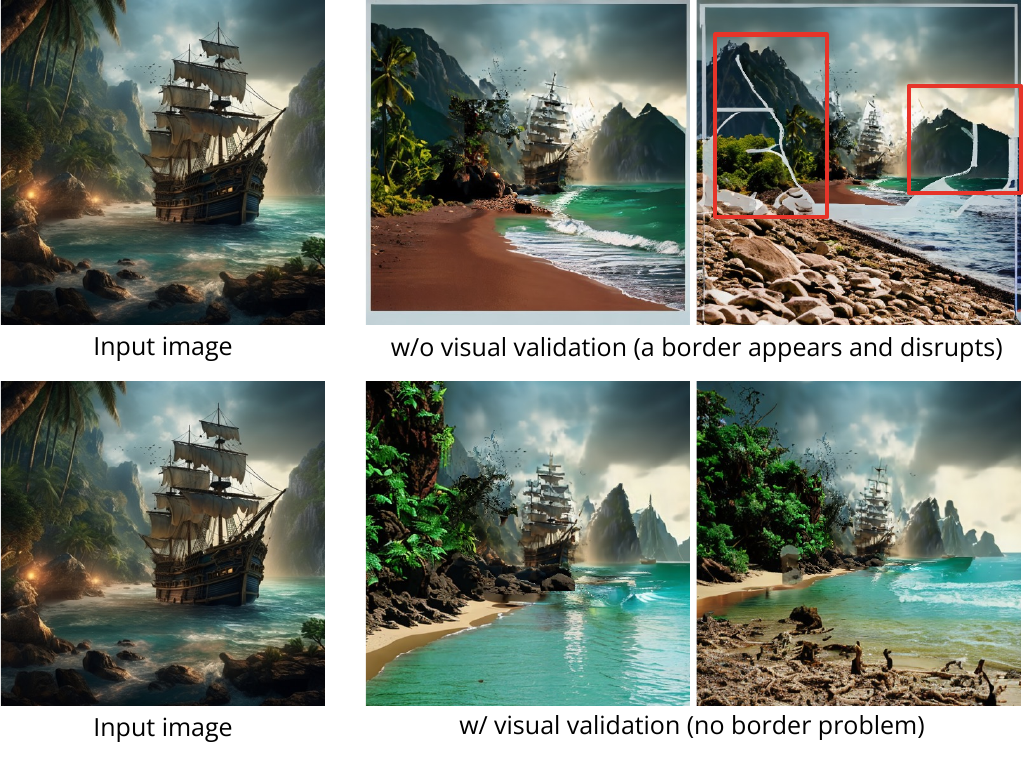}
    \vspace{-0.4cm}
    \caption{\textbf{Ablation comparison on using a VLM} to visually check if borders appear. Without visual validation, borders often appear and it disrupts the following scenes. Red boxes highlight the border disruptions.}
    \label{fig:ablation_border}
\end{figure}
\myparagraph{Visual validation by a VLM.} As mentioned in \sect{\ref{sec:renderer}}, the painting frame and photo border can be a strong disruption. We show an example in \fig{\ref{fig:ablation_border}}, where the photo border appears and disrupts the next generated scene. The visual validation can effectively detect borders and launch a re-generation process to handle it.

\begin{figure}
    \centering
    \includegraphics[width=\linewidth]{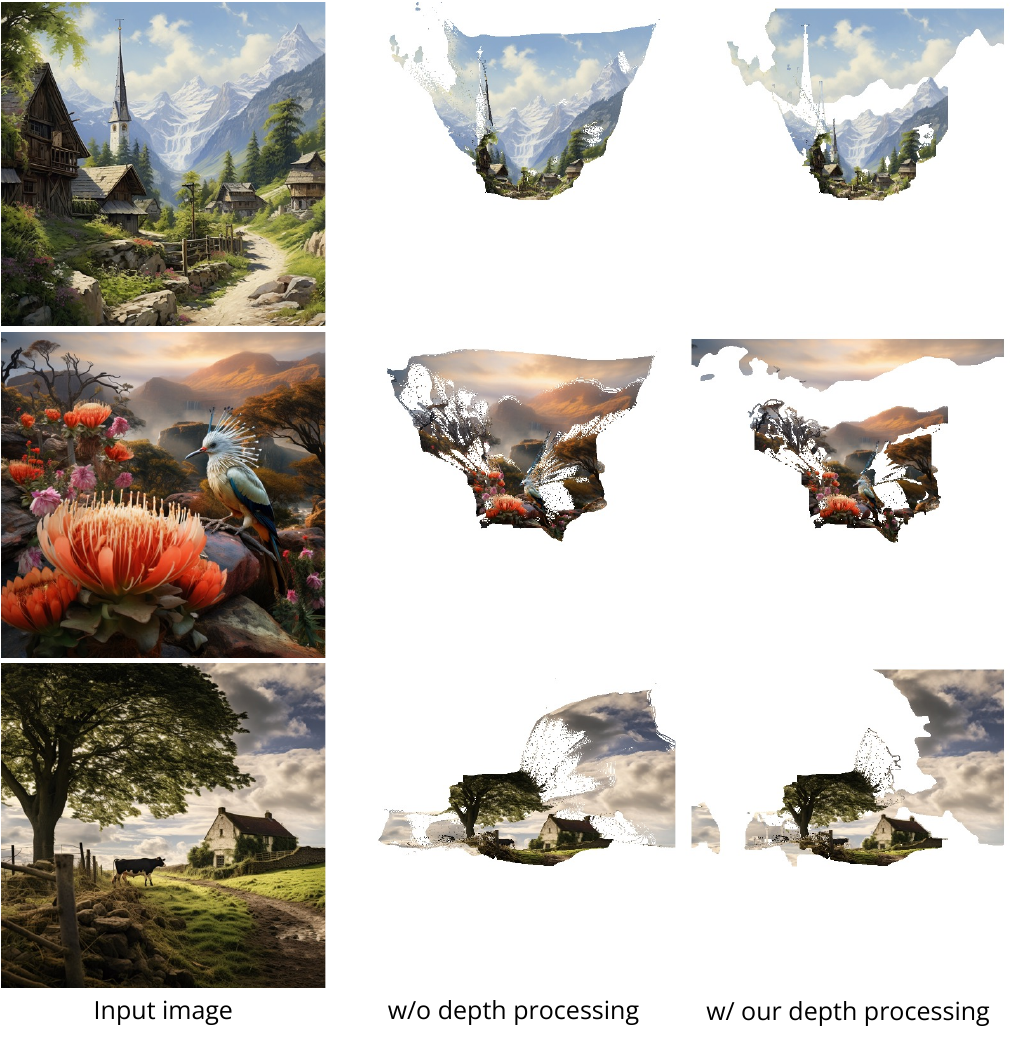}
    \vspace{-0.4cm}
    \caption{\textbf{Ablation comparison on our depth processing.} Zoom in to see more details. Without our depth processing, the rendered partial image demonstrates strong distortions. For examples, see the sky (bottom example), the bird (middle example), and the tower (top example).}
    \label{fig:ablation_depth}
\end{figure}
\myparagraph{Depth processing.} Our depth processing is essential in generating geometrically coherent scenes. In \fig{\ref{fig:ablation_depth}}, we show a visual comparison, where we show the rendered partial images using the original estimated depth and using our processed depth. Without our depth processing, the partial rendered images show strong distortions due to depth inaccuracy in object boundary, sky, and distant areas. For example, see the distortions in the sky (bottom example), the bird (middle example), and the tower (top example) in \fig{\ref{fig:ablation_depth}}.
\section{Details on the LLM and VLM}\label{sec:prompts}

We use GPT-4 for generating scene descriptions. Specifically, we use the following prompt $\mathcal{J}$ for the system call:

\textit{``You are an intelligent scene generator. Imaging you are flying through a scene or a sequence of scenes, and there are 3 most significant common entities in each scene. Please tell me what sequentially next scene would you likely to see? You need to generate the scene name and the 3 most common entities in the scene. The scenes are sequentially interconnected, and the entities within the scenes are adapted to match and fit with the scenes. You also have to generate a brief background prompt about 50 words describing the scene. You should not mention the entities in the background prompt. If needed, you can make reasonable guesses.''}

The input is the scene description memory $\mathcal{M}$ which is a collection of past and current scene descriptions $\mathcal{M}_i = \{\mathcal{S}_0, \cdots, \mathcal{S}_i\}$ as defined in \eqn{\ref{eqn:memory}}. In particular, $\mathcal{S}_i = \{S, O_i, B_i\}$ where $S$ denotes a style, $O_i$ denotes the object description, and $B_i$ denotes the background description. We use a lexical category filter to extract nouns and adjectives. An actual scene description $\mathcal{S}_i$ looks like (the following is the actual text prompts of the ``girl in wonderland'' example in \fig{\ref{fig:teaser}}):

\begin{itemize}
    \item \textit{Scene 1: $\{$Background: Alice in the wonderland. Entities: [' Alice', ' flowers', ' cat']; Style: Monet painting$\}$}
    \item \textit{Scene 2: $\{$Background: luminous painting, way, vibrant bizarre croquet field, player, flamingo, mallet, Hedgehogs, balls, croquet, balls, life, domineering presence, atmosphere. Entities: ['flamingos', 'hedgehogs', 'The Queen of Hearts']; Style: Monet painting$\}$}
    \item \textit{Scene 3: $\{$Background: scene, bizarre tea, party, great elm, tree, eccentric gentleman, top hat, presides, celebration, jittery hare, sleepy rodent, Cups, plates, assorted pastries, ancient misshapen, table, atmosphere, chaotic random bouts, nonsensical poetry, riddles. Entities: ['Mad Hatter', 'March Hare', 'Dormouse']; Style: Monet painting$\}$}
    \item \textit{Scene 4: $\{$Background: impressionist strokes, endless checkerboard, landscape, animate chess, pieces, rules, game, sense, tension, serene, countryside, ambiance, trees, strange fruits, flowers. Entities: ['White Queen', 'Red Queen', 'Pawn']; Style: Monet painting$\}$}
\end{itemize}

We use GPT-4V as the VLM for visual validation. The most significant unwanted effects are the painting frame and photo border that appear along some of the four boundaries of the outpainted image. We use the following system call:

\textit{``Along the four borders of this image, is there anything that looks like thin border, thin stripe, photograph border, painting border, or painting frame? Please look very closely to the four edges and try hard, because the borders are very slim and you may easily overlook them. If you are not sure, then please say yes.''}

We also use similar prompt for detecting out-of-focus blurry objects. If GPT-4V fails due to network connection or other practical reasons, we instead generate an $560\times 560$ image and then center-crop it to bypass the frame problem. This is not as good as using visual validation, because it can lead to cropped partial foreground objects such as a half of a person, and the partial objects can become floaters due to low depth values when we move camera to generate new scenes.
\section{Details on Human Preference Evaluation}\label{sec:human_study}

We use Prolific\footnote{\url{https://www.prolific.com/}} to recruit participants for the human preference evaluation. We use Google forms to present the survey. The survey is fully anonymized for both the participants and the host. We attach the anonymous survey link in the
footnote\footnote{Comparison to InfiniteNature-Zero: \url{https://forms.gle/mKxyJUT3qZLs2f8h9}; Comparison to SceneScape: \url{https://forms.gle/pt7NBj73Fnd5apjM8}. Google forms require signing in to participate, but it does not record participant's identity.} for reference.

\end{document}